\documentclass[a4paper,fleqn]{cas-sc}
\usepackage[numbers,sort&compress]{natbib}
\usepackage{mathtools,amssymb,bm}
\usepackage{amsthm}
\usepackage{booktabs,array,tabularx,microtype,url}
\usepackage{graphicx,tikz,placeins}
\usepackage{trimclip} %
\usetikzlibrary{arrows.meta,positioning}
\hypersetup{hidelinks}
\newcommand{\R}{\mathbb R}
\newcommand{\SE}{\mathrm{SE}}
\newcommand{\SO}{\mathrm{SO}}

\newcommand{\rot}{\boldsymbol{\mathcal R}}
\newcommand{\Rp}{\widetilde{\boldsymbol R}_{p}}

\newcommand{\K}{\boldsymbol K}
\newcommand{\Dzero}{(D\widetilde{\boldsymbol\psi})_0}
\newcommand{\Dstar}{(D\widetilde{\boldsymbol\psi})^*}
\newcommand{\Dhat}{\widehat{D\widetilde{\boldsymbol\psi}}}
\newcommand{\SX}{\boldsymbol S_X}
\newcommand{\Manc}{\boldsymbol M_{\mathrm{anc}}}
\newcommand{\Janc}{\boldsymbol J_{\mathrm{anc}}}
\DeclareMathOperator{\diag}{diag}
\DeclareMathOperator{\col}{col}
\DeclareMathOperator{\blkdiag}{blkdiag}
\DeclareMathOperator{\Ad}{Ad}
\DeclareMathOperator{\ad}{ad}
\DeclareMathOperator{\Log}{Log}

\newtheorem{lemma}{Lemma}
\newtheorem{theorem}{Theorem}
\newtheorem{corollary}{Corollary}
\newtheorem{proposition}{Proposition}
\theoremstyle{definition}\newtheorem{assumption}{Assumption}
\theoremstyle{remark}
\ExplSyntaxOn
\cs_set:Npn \__first_footerline:
 {
  \group_begin:
  \small\sffamily
  \ifnum\theblind>0\relax
  \else
    \tl_if_blank:eF {\__short_authors:} {\__short_authors: :~}
  \fi
  {\rmfamily\itshape Preprint~submitted~to~Elsevier}
  \group_end:
 }
\ExplSyntaxOff
\makeatletter
\ExplSyntaxOn
\exp_args:Nc \patchcmd { environment~Abstract~code }
 { \hbox_to_wd:nn { \z@ } { \box \g_stm_key_box } }
 { \hbox_to_wd:nn { \z@ } { \box \g_stm_key_box \hss } }
 { }
 { \PackageError{pathassembly}{CAS~keyword~overlay~patch~failed}{Check~the~CAS~class~version.} }
\ExplSyntaxOff
\makeatother
\begin{document}
\let\WriteBookmarks\relax
\shorttitle{Reconstruction-Error Propagation and Task-Jacobian Accuracy}
\shortauthors{Mohammad Dastranj and Jouni Mattila}
\title[mode=title]{Modular Reconstruction-Error Propagation and Task-Jacobian Accuracy Bounds for Path-Assembled Closed-Chain Mechanism Models}
\author[1]{Mohammad Dastranj}[orcid=0009-0005-7116-4496]
\cormark[1]
\ead{mohammad.dastranj@tuni.fi}
\author[1]{Jouni Mattila}[orcid=0000-0003-1799-4323]
\ead{jouni.mattila@tuni.fi}
\affiliation[1]{organization={Unit of Automation Technology and Mechanical Engineering, Faculty of Engineering and Natural Sciences, Tampere University},
  postcodesep={},postcode={33720},city={Tampere},country={Finland}}
\cortext[1]{Corresponding author}
\begin{abstract}
Numerical reconstruction of closed-chain mechanisms introduces errors in passive coordinates and in the differential maps used to assemble task kinematics. This paper develops a conditional accuracy analysis for path-assembled mechanism models that connects module-local closure residuals to task-level error bounds. Local inverse and derivative bounds provide passive-coordinate and differential-map allowances, including corrections associated with computed centers. These allowances are propagated through ordered transformation paths while preserving module support, exact independent-coordinate rows, and nonlinear cross terms. With declared coordinate and task scales, the resulting dimensionless bound on the normalized reduced task-Jacobian error gives sufficient module-specific reconstruction requirements for a prescribed task-level budget. Supporting static end-point position and orientation bounds and same-input instantaneous task-rate allowances provide physical interpretations of the reconstruction accuracy. In a fixed-configuration assessment on a closed-chain robot, nominal-double Jacobian and pose discrepancies from refined numerical references remain within the calculated allowances, while a directional finite-difference check supports the analytical Jacobian. Structured propagation yields smaller bounds than the shared-input normwise comparator and accepts more of the available module-budget combinations at an illustrative threshold. Contribution diagnostics identify the differential-map term as dominant in the conservative Jacobian allowances, with positive sufficient task-rank margins in all evaluated trials. The measured cost of bound evaluation exceeds that of reconstruction. The analysis applies on a regular local branch with explicit neighborhood, logarithm-domain, and complete-closure compatibility conditions.
\end{abstract}
\maketitle
\raggedbottom %

\section{Introduction}
\label{sec:introduction}
Closed-chain mechanisms couple independently prescribed joint coordinates to passive coordinates through nonlinear loop-closure equations. Once a regular configuration is reconstructed, implicit differentiation provides the active-to-passive map used to reduce the mechanism kinematics. A finite closure tolerance therefore affects two connected quantities: the reconstructed configuration and the differential map evaluated there. Their errors subsequently enter the end-point pose and reduced task Jacobian. Relating a residual tolerance to these errors requires information about local conditioning and derivative variation, as well as the way module-level quantities enter the assembled model.

The accuracy required in kinematic reconstruction is ultimately governed by the admissible error at the task level. Further closure correction beyond what is necessary to satisfy this requirement may increase computational effort without providing a meaningful engineering benefit. Conversely, relaxing reconstruction tolerances without accounting for the sensitivity of the assembled kinematics may compromise the required accuracy. Establishing a quantitative relationship between local reconstruction tolerances and the resulting task-kinematic errors therefore provides a basis for selecting accuracy requirements commensurate with the application. Within the path assembly framework, such a relationship also enables these requirements to be differentiated among closed-chain modules according to their influence on the assembled kinematics.

This focus on numerical accuracy complements research on the identification and compensation of physical model errors. In overconstrained and parallel mechanisms, calibration models account for geometric deviations, internal deformation, and passive-joint effects~\cite{ye2023overconstrained,wu2024elastogeometric,yuan2024passive}. Recent calibration approaches also address weighted parameter identification~\cite{gu2025weighted} and left-invariant pose-error representations~\cite{huang2025leftinvariant}, configuration-dependent kinematic parameters~\cite{lu2025configuration}, and probabilistic end-point error distributions~\cite{gerstewitz2026distributions}. These studies improve the correspondence between a mechanism and its model. The present analysis instead quantifies the numerical error introduced when the closure equations of a fixed nominal model are solved to finite accuracy.

Several complementary approaches address the underlying closure problem. Monodromy-based continuation has been used to compute forward-kinematic equilibria of cable-driven parallel robots with sagging cables~\cite{baskar2024monodromy}, while symbolic methods and continuation combined with interval analysis support the identification of singularity-free regions under design uncertainty~\cite{le2025certified}. Recent coordinate-partitioning studies compare dependent-coordinate updates in terms of constraint violation, kinematic accuracy, and computational cost within multibody simulation~\cite{zhang2026partitioning}. For single-degree-of-freedom closed-loop mechanisms, $C^2$ Hermite interpolation has also been used to reconstruct intermediate configurations and improve the smoothness of angular velocities and accelerations~\cite{capette2026hermite}. These developments motivate an explicit treatment of reconstruction accuracy and its effect on differential kinematics.

Modular representations provide a natural setting for this treatment. Local loop-closure functions and their derivatives can be composed to model series-parallel hybrid mechanisms~\cite{kumar2025modular}. In the path assembly framework~\cite{dastranj2026}, closed-chain modules are represented by ordered transformation paths and selected independent closure equations. Defect homotopy acquires a physical branch under feasibility and regularity conditions, and predictor--corrector reconstruction supplies nearby passive configurations. The resulting local differential maps connect module reconstruction to the global reduced kinematics. This representation also underlies a recent inverse-dynamics and wrench-recovery formulation~\cite{dastranj2026dynamics}; the present paper addresses numerical kinematic accuracy within the same framework, retaining the original branch and complete-closure assumptions.

The accuracy of a derivative obtained from an approximate implicit solution is a broader numerical-analysis question. Inexact-Newton forcing theory controls the accuracy of linear solves within nonlinear iterations~\cite{eisenstat1996forcing}. In bilevel learning with smooth, strongly convex lower-level problems, a priori and computable a posteriori bounds quantify how an approximate inner solution and an inexact implicit linear solve affect derivatives of the outer objective~\cite{ehrhardt2024inexact}. These results provide a relevant antecedent for separating reconstruction error from differential-solve error. Contraction arguments~\cite{ortega1970iterative} and radius-admissibility conditions used in radii-polynomial methods~\cite{hungria2016radii} provide antecedents for the local analysis on a prescribed regular branch; singular-value perturbation estimates~\cite{stewart1990svd} support the separate task-rank assessment.

The question addressed in this paper is how these local error allowances propagate through the path assembly framework, and which module-specific allowances are sufficient for a prescribed task-Jacobian budget. The propagation must account for shared independent coordinates, exact identity rows, and nonlinear cross terms in ordered transformation products. It must also distinguish the reconstructed coordinates from the differential maps stored at that center. Since joint coordinates and logarithmic closure residuals can contain both angular and translational quantities, fixed input and output scales are essential to interpreting the resulting norms~\cite{doty1993generalized,lynch2017modern}. Accordingly, the primary criterion is a dimensionless normalized task-Jacobian operator error, which bounds the effect of reconstruction on the differential map for every input direction under the declared scaling. Static position and orientation allowances provide complementary physical-unit interpretations.

The first contribution is a set of conditional local bounds for passive-coordinate reconstruction and the active-to-passive differential map. Selected-closure residual allowances, local inverse bounds, and derivative majorants are combined with the corrections required at computed centers. The regularity, neighborhood, logarithm-domain, and complete-closure compatibility conditions remain explicit in the construction.

The second contribution is structure-preserving propagation of these module-local allowances to the normalized reduced task Jacobian. The construction retains module support, exact independent-coordinate rows, and finite perturbation cross terms. The same path enclosure supplies supporting static end-point position and orientation bounds, together with a same-input instantaneous task-rate allowance.

The third contribution is a sufficient condition on module-specific reconstruction accuracies for meeting a prescribed normalized task-Jacobian budget. It connects local reconstruction requirements to a common task-level criterion, allowing combinations of module allowances to be assessed under their respective validity conditions.

The numerical assessment uses nine accepted fixed configurations of an actual closed-chain robot and 27 global reconstruction trials. It examines the calculated allowances against numerical references, compares structured and shared-input normwise bounds, and reports module-budget availability and computational overhead. End-point pose comparisons, directional-Jacobian checks, and task-rank margins provide complementary assessments. Contribution diagnostics examine the origin of the conservative allowances. All reported numerical evaluations use ordinary double precision under the conventions stated in Section~\ref{sec:nominal}.

The remainder of the paper is organized as follows. Section~\ref{sec:nominal} introduces the nominal model and normalization. Section~\ref{sec:local} establishes the local reconstruction and differential-map bounds, and Section~\ref{sec:task} develops their structure-preserving task propagation. Section~\ref{sec:feasibility} presents the sufficient accuracy requirements and numerical assessment, followed by the conclusions in Section~\ref{sec:conclusion}.

\section{Nominal Model and Conventions}
\label{sec:nominal}
\subsection{Coordinates, closure, and branch meaning}
The nominal configuration is described by a global coordinate vector $\boldsymbol q\in\R^{n_q}$, comprising the prescribed independent coordinates $\boldsymbol a\in\R^{n_a}$ and the passive coordinates $\boldsymbol p_m$ of each closed-chain module $m$. In this partition, each passive coordinate belongs to one module, although different modules are allowed to depend on the same independent coordinates. Directly coupled passive coordinates are represented within a single module, and all coordinates outside the passive sets are prescribed. For each module, $\boldsymbol a_m$ denotes the subset of independent coordinates on which its closure depends.

The transform $\boldsymbol T_{ij}$ maps frame-$j$ components into frame $i$. Each closure element $\ell$ compares two ordered paths $\boldsymbol T_\ell^{\langle+\rangle}$ and $\boldsymbol T_\ell^{\langle-\rangle}$ with common initial and terminal frames. Their relative transform is $\boldsymbol\Delta_\ell=(\boldsymbol T_\ell^{\langle-\rangle})^{-1}\boldsymbol T_\ell^{\langle+\rangle}$, with angular-first logarithmic coordinates $\boldsymbol\rho_\ell=\Log(\boldsymbol\Delta_\ell)^\vee$. The selected closure equations for module $m$ are
\begin{equation}
 \boldsymbol c_m=\boldsymbol S_m\col(\boldsymbol\rho_\ell).
 \label{eq:selected-constraints}
\end{equation}
The fixed row selector $\boldsymbol S_m$ retains exactly $n_{m,p}=\dim\boldsymbol p_m$ scalar equations. The Jacobian blocks $\boldsymbol R_{m,p}=D_{\boldsymbol p_m}\boldsymbol c_m$ and $\boldsymbol R_{m,a}=D_{\boldsymbol a_m}\boldsymbol c_m$ collect their derivatives with respect to the passive and independent module coordinates, respectively. Here $D_{\boldsymbol x}$ denotes differentiation with respect to $\boldsymbol x$, while the coordinate differentials satisfy $\mathrm d\boldsymbol c_m=\boldsymbol R_{m,p}\,\mathrm d\boldsymbol p_m+\boldsymbol R_{m,a}\,\mathrm d\boldsymbol a_m$.

\begin{assumption}[Local nominal branch and domains]
\label{ass:branch}
Defect-homotopy acquisition starts from the exact artificial solution defined by the initial passive-coordinate estimate and follows the branch selected by this estimate and the prescribed passive bounds~\cite{dastranj2026}. This continuation is assumed to reach the intended physical closure solution along a feasible, regular path within those bounds and the adopted logarithm domain. The nominal factors are rigid transforms or constant-twist exponentials, and the physical residual rows selected by rank-revealing analysis remain fixed on each analysis neighborhood. The selected closure is $C^2$ on an open neighborhood of each admissible closed passive box, with a single smooth logarithm branch and a nonsingular passive Jacobian at its center. The inherited branch and complete-closure compatibility conditions identify the selected local root with the intended physical configuration.
\end{assumption}

An asterisk denotes an exact nominal-model quantity on the physical branch specified by Assumption~\ref{ass:branch}. Thus $\boldsymbol p_m^*=\boldsymbol\psi_m(\boldsymbol a_m)$ and $\boldsymbol q^*=\boldsymbol\Psi(\boldsymbol a)$ are the exact passive and global configurations, respectively. Implicit differentiation yields $D\boldsymbol\psi_m=-\boldsymbol R_{m,p}^{-1}\boldsymbol R_{m,a}$, where $D\boldsymbol\psi_m$ is the active-to-passive Jacobian and the residual blocks are evaluated on this branch. Let $\boldsymbol E_a$ insert the global active rows and $\boldsymbol E_{p,m}$ the passive rows of module $m$, while $\boldsymbol L_m$ selects $\boldsymbol a_m$ from $\boldsymbol a$. The resulting global differential map is
\begin{equation}
 \boldsymbol M^*=D\boldsymbol\Psi(\boldsymbol a)
 =\boldsymbol E_a+\sum_m\boldsymbol E_{p,m}D\boldsymbol\psi_m\boldsymbol L_m.
 \label{eq:global-reconstruction}
\end{equation}
The insertion and selection matrices are fixed zero--one matrices determined by the coordinate structure.

These exact quantities provide the references for the error analysis of a computed reconstruction and its differential map. Section~\ref{sec:local} develops local residual and derivative conditions that ensure a unique selected-closure root near the computed configuration and provide the corresponding error bounds. The branch and complete-closure compatibility conditions of Assumption~\ref{ass:branch} identify this root with the intended physical configuration. Numerical acceptance additionally requires checks of the complete raw closure residual and local rank.

\subsection{Scales and exact versus computed objects}
\label{subsec:numerical_conventions}
Within each analysis neighborhood, the coordinate and residual normalizations are specified by fixed positive diagonal matrices $\boldsymbol\Lambda_{m,a}$, $\boldsymbol\Lambda_{m,p}$, and $\boldsymbol\Lambda_{m,c}$. Relative to fixed offsets, the independent coordinates are written as $\boldsymbol a_m=\boldsymbol a_{m,\mathrm{off}}+\boldsymbol\Lambda_{m,a}\widetilde{\boldsymbol a}_m$, and the passive coordinates as $\boldsymbol p_m=\boldsymbol p_{m,\mathrm{off}}+\boldsymbol\Lambda_{m,p}\widetilde{\boldsymbol p}_m$. The selected residual in these coordinates is normalized as $\widetilde{\boldsymbol c}_m=\boldsymbol\Lambda_{m,c}^{-1}\boldsymbol c_m$. Its Jacobian blocks with respect to the normalized module coordinates are therefore
\begin{equation}
 \widetilde{\boldsymbol R}_{m,b}
 =\boldsymbol\Lambda_{m,c}^{-1}\boldsymbol R_{m,b}\boldsymbol\Lambda_{m,b},
 \qquad b\in\{a,p\}.
 \label{eq:normalization}
\end{equation}
The same coordinate transformation gives the normalized active-to-passive differential map as
\begin{equation}
 D\widetilde{\boldsymbol\psi}_m
 =\boldsymbol\Lambda_{m,p}^{-1}D\boldsymbol\psi_m\boldsymbol\Lambda_{m,a}.
 \label{eq:normalized_differential_map}
\end{equation}
These transformations account for the scales of the coordinate inputs and the residual or passive-coordinate outputs. Scaling the coordinates together with the residuals gives a consistent interpretation of mixed-coordinate norms, in accordance with the role of metrics in robot inverse kinematics~\cite{doty1993generalized}. For a consistent comparison, coordinate scales are defined from predeclared operating-interval widths or justified positive characteristic excursions, with all module scales extracted from one global scale vector. The scales are fixed before the bounds are evaluated. Across configurations, the coordinate and task scales are kept common for comparison, whereas the residual scales may be constructed locally and then held fixed on the corresponding analysis neighborhood, as in Subsection~\ref{subsec:numerical_setup}.

The subscript $0$ denotes an exact nominal function evaluated at the declared center, a hat denotes a computed quantity, and $\mathit{ref}$ identifies a numerical reference. A center specified in floating-point arithmetic still defines a point at which the nominal functions have exact values. For comparison with the exact global differential map $\boldsymbol M^*$ in~\eqref{eq:global-reconstruction}, the \emph{map anchor} $\Manc$ is a fixed matrix assembled from exact or computed local differential maps at the declared center, retaining the exact active-coordinate identity rows. The nonnegative matrix $\boldsymbol B_M$ provides the associated entrywise error allowance, $|\boldsymbol M^*-\Manc|\leq\boldsymbol B_M$. Section~\ref{sec:task} constructs this allowance from the module-local bounds, including center-map computation errors when computed local maps are used. Absolute values and array inequalities are componentwise; matrix norms are induced unless specified otherwise.

All theorem inequalities are stated over the real numbers, whereas the numerical study evaluates the corresponding bounds in ordinary double precision without outward rounding or certified evaluation-error enclosures. The resulting values are therefore interpreted as nominal evaluations under the stated assumptions, with numerical-reference comparisons used to assess their consistency. A verified finite-precision interpretation would additionally require justified enclosures of function-evaluation errors and rounding effects; Appendix~\ref{app:numerical} sets out the corresponding finite-evaluation terms and numerical acceptance conditions.

\section{Local Reconstruction and Differential Accuracy}
\label{sec:local}
\subsection{Local closure reconstruction}
\label{subsec:local_closure_reconstruction}
The local analysis concerns one module at fixed independent coordinates, with the module subscript omitted for readability. The reconstructed normalized passive coordinates $\widetilde{\boldsymbol p}_0$ define the center of the closed box $\mathcal B(\varsigma)=\{\widetilde{\boldsymbol p}:\|\widetilde{\boldsymbol p}-\widetilde{\boldsymbol p}_0\|_\infty\leq \varsigma\}$, where the scalar $\varsigma>0$ is its radius in the infinity norm. The box lies within the admissible passive region and the smooth-logarithm domain of Assumption~\ref{ass:branch}.

With the independent coordinates held fixed, $\widetilde{\boldsymbol c}(\widetilde{\boldsymbol p})$ denotes the selected normalized closure residual and $\Rp(\widetilde{\boldsymbol p})=D_{\widetilde{\boldsymbol p}}\widetilde{\boldsymbol c}(\widetilde{\boldsymbol p})$ its passive-coordinate Jacobian. Their exact nominal values at the center are $\widetilde{\boldsymbol c}_0$ and $\widetilde{\boldsymbol R}_{p,0}$. For a nonnegative componentwise residual allowance $\overline{\boldsymbol c}_0$, the inverse center Jacobian and the associated preconditioned residual magnitudes are
\begin{equation}
 \K=\widetilde{\boldsymbol R}_{p,0}^{-1},\qquad
 Y_{\mathrm{obs}}=\|\K\widetilde{\boldsymbol c}_0\|_\infty,\qquad Y=\||\K|\overline{\boldsymbol c}_0\|_\infty,\qquad
 |\widetilde{\boldsymbol c}_0|\leq\overline{\boldsymbol c}_0.
 \label{eq:newton_budget}
\end{equation}
Here $\K$ is the exact inverse of the normalized passive Jacobian at the center, $Y_{\mathrm{obs}}$ is the norm of the Newton correction associated with the particular exact nominal residual, and $Y$ is its upper bound obtained from $\overline{\boldsymbol c}_0$, so that $Y_{\mathrm{obs}}\leq Y$. The residual allowance may represent a justified stopping budget or a sharper bound on the center residual. Its construction is recorded, including the evaluation-uncertainty allowance needed when it is inferred from a computed residual, as described in Appendix~\ref{app:numerical}.

Variation of the passive Jacobian across the box is described by its directional derivative $D\Rp(\widetilde{\boldsymbol p})[\boldsymbol h]$, where $\boldsymbol h$ is a normalized passive-coordinate direction. A nonnegative curvature allowance $\Gamma_{pp}(\varsigma)$ satisfies the uniform bound
\begin{equation}
 \|\K D\Rp(\widetilde{\boldsymbol p})[\boldsymbol h]\|_\infty
 \leq\Gamma_{pp}(\varsigma)\|\boldsymbol h\|_\infty.
 \label{eq:preconditioned_closure_curvature}
\end{equation}
This inequality holds for every $\widetilde{\boldsymbol p}\in\mathcal B(\varsigma)$ and every direction $\boldsymbol h$. Appendix~\ref{app:path_majorants} constructs such preconditioned curvature bounds from the declared paths.

\begin{lemma}[Local closure reconstruction]
\label{lem:local_reconstruction}
Under the stated residual-budget, domain, and curvature conditions, the following inequalities on the radius $\varsigma$ are sufficient for local reconstruction:
\begin{equation}
 Y+\tfrac12\Gamma_{pp}(\varsigma)\varsigma^2\leq \varsigma,
 \qquad\mu:=\Gamma_{pp}(\varsigma)\varsigma<1.
 \label{eq:reconstruction_radius_conditions}
\end{equation}
They ensure a unique selected-closure root $\widetilde{\boldsymbol p}^*$ in $\mathcal B(\varsigma)$. The corresponding reconstruction and inverse-Jacobian bounds are
\begin{equation}
 \|\widetilde{\boldsymbol p}^*-\widetilde{\boldsymbol p}_0\|_\infty
 \leq\varepsilon_p:=\varsigma,\qquad \|\Rp(\widetilde{\boldsymbol p})^{-1}\|_\infty
 \leq\frac{\|\K\|_\infty}{1-\mu},\qquad \widetilde{\boldsymbol p}\in\mathcal B(\varsigma).
 \label{eq:normalized_reconstruction_bound}
\end{equation}
Here $\mu$ bounds the contraction factor on the box, and $\varepsilon_p$ is the normalized passive-coordinate reconstruction-error allowance.
\end{lemma}
\begin{proof}
The map $\mathcal T(\widetilde{\boldsymbol p})=\widetilde{\boldsymbol p}-\K\widetilde{\boldsymbol c}(\widetilde{\boldsymbol p})$ uses the Jacobian inverse fixed at the center. Since $\K\widetilde{\boldsymbol R}_{p,0}=\boldsymbol I$, where $\boldsymbol I$ is the identity on the passive-coordinate space, the derivative of $\mathcal T$ vanishes at $\widetilde{\boldsymbol p}_0$. Integration of~\eqref{eq:preconditioned_closure_curvature} along the segment from the center to any point in the box therefore gives $\|D\mathcal T(\widetilde{\boldsymbol p})\|_\infty\leq\Gamma_{pp}(\varsigma)\|\widetilde{\boldsymbol p}-\widetilde{\boldsymbol p}_0\|_\infty\leq\mu<1$.

For a displacement $\boldsymbol h$ with $\|\boldsymbol h\|_\infty\leq \varsigma$, Taylor's formula about the center bounds $\|\mathcal T(\widetilde{\boldsymbol p}_0+\boldsymbol h)-\widetilde{\boldsymbol p}_0\|_\infty$ by $Y+\Gamma_{pp}(\varsigma)\|\boldsymbol h\|_\infty^2/2$. The first inequality in~\eqref{eq:reconstruction_radius_conditions} makes this bound no larger than $\varsigma$, so $\mathcal T$ maps the closed box into itself. The box is complete, and the contraction theorem~\cite{ortega1970iterative} consequently provides a unique fixed point. Nonsingularity of $\K$ identifies this fixed point with the selected-closure root $\widetilde{\boldsymbol p}^*$ and gives the stated reconstruction bound.

Throughout the same box, the identity $\K\Rp=\boldsymbol I+\K(\Rp-\widetilde{\boldsymbol R}_{p,0})$ expresses the preconditioned Jacobian as an identity plus a perturbation of norm at most $\mu$. Its Neumann-series inverse is therefore bounded by $(1-\mu)^{-1}$. Multiplication by $\K$ yields the inverse-Jacobian bound in~\eqref{eq:normalized_reconstruction_bound}.
\end{proof}
The radius-search iteration $\varsigma_{i+1}=Y+\Gamma_{pp}(\varsigma_i)\varsigma_i^2/2$, where $i$ indexes successive candidate radii, provides a numerical procedure for finding a box satisfying~\eqref{eq:reconstruction_radius_conditions}. Once such a box is established, the particular center residual gives the optional sharper root-distance allowance $\min\{\varsigma,Y_{\mathrm{obs}}/(1-\mu)\}$ in place of $\varepsilon_p$, while the original box and its uniform constants are retained. Any modification of the inverse factor is supported by the corresponding valid derivative bound.

\subsection{Differential-map sensitivity}
\label{subsec:differential_map_sensitivity}
The exact nominal differential map at the center is $\Dzero=-\K\widetilde{\boldsymbol R}_{a,0}$, where $\widetilde{\boldsymbol R}_{a,0}$ is the normalized active-coordinate residual-Jacobian block evaluated there. The map $\Dstar$ is evaluated at the selected root $\widetilde{\boldsymbol p}^*$. Their difference depends on the variation of both residual-Jacobian blocks over the same passive box.

For $b\in\{a,p\}$, the nonnegative matrices $\boldsymbol{\mathcal H}^{pb}_{j}$ bound $|\partial\widetilde{\boldsymbol R}_{b}/\partial\widetilde p_j|$ componentwise throughout $\mathcal B(\varsigma)$. Here $j$ ranges over the normalized passive coordinates, and $b$ identifies the active or passive Jacobian block; each bounding matrix has the same dimensions as that block. These bounds on second derivatives of the selected residual give the preconditioned sensitivity allowances
\begin{equation}
 \Gamma_{pa}=\sum_j\||\K|\boldsymbol{\mathcal H}^{pa}_{j}\|_\infty,\qquad \Gamma_{pD}=\sum_j\||\K|\boldsymbol{\mathcal H}^{pp}_{j}|\Dzero|\|_\infty.
 \label{eq:differential_curvature_allowances}
\end{equation}
The quantity $\Gamma_{pa}$ bounds variation of the preconditioned active block with passive-coordinate displacement. The quantity $\Gamma_{pD}$ bounds variation of the preconditioned passive block acting on the center differential map; this product is formed before the column information is reduced to a scalar norm.

\begin{lemma}[Differential-map error]
\label{lem:differential_map_error}
Under Lemma~\ref{lem:local_reconstruction} and the stated Hessian bounds, the normalized differential-map error satisfies
\begin{equation}
 \|\Dstar-\Dzero\|_\infty
 \leq\frac{\varepsilon_p(\Gamma_{pa}+\Gamma_{pD})}{1-\mu}
 =:\varepsilon_D.
 \label{eq:differential_map_error_bound}
\end{equation}
\end{lemma}
\begin{proof}
The block perturbations are $\Delta\widetilde{\boldsymbol R}_b=\widetilde{\boldsymbol R}_b^*-\widetilde{\boldsymbol R}_{b,0}$ for $b\in\{a,p\}$, with $\widetilde{\boldsymbol R}_b^*$ evaluated at $\widetilde{\boldsymbol p}^*$. Subtracting the center relation from the implicit relation at the root and solving for the map difference gives
\begin{equation}
 \Dstar-\Dzero=-(\K\Rp^*)^{-1}\K
 [\Delta\widetilde{\boldsymbol R}_a+\Delta\widetilde{\boldsymbol R}_p\Dzero].
 \label{eq:differential_perturbation_identity}
\end{equation}
Integration along the segment from $\widetilde{\boldsymbol p}_0$ to $\widetilde{\boldsymbol p}^*$ bounds $\|\K\Delta\widetilde{\boldsymbol R}_a\|_\infty$ by $\varepsilon_p\Gamma_{pa}$ and $\|\K\Delta\widetilde{\boldsymbol R}_p\Dzero\|_\infty$ by $\varepsilon_p\Gamma_{pD}$. Lemma~\ref{lem:local_reconstruction} supplies the inverse-factor bound $\|(\K\Rp^*)^{-1}\|_\infty\leq(1-\mu)^{-1}$. Applying these bounds to~\eqref{eq:differential_perturbation_identity} yields~\eqref{eq:differential_map_error_bound}, with $\varepsilon_D$ denoting the resulting scalar error allowance.
\end{proof}
A simpler valid normwise numerator for~\eqref{eq:differential_map_error_bound} is $\varepsilon_p\Gamma_{pa}+\mu\|\Dzero\|_\infty$. Both numerators use the same root box and inverse factor, allowing their comparison with the other local bounds held fixed.

For a stored computed normalized differential map $\Dhat$ at the center, the matrix $\boldsymbol r_D$ measures its residual in the linear relation formed from the exact nominal center blocks. The resulting correction gives
\begin{equation}
 \boldsymbol r_D=\widetilde{\boldsymbol R}_{p,0}\Dhat+\widetilde{\boldsymbol R}_{a,0},\qquad \|\Dstar-\Dhat\|_\infty\leq\varepsilon_D+\|\K\boldsymbol r_D\|_\infty.
 \label{eq:computed_differential_map_bound}
\end{equation}
The identity $\Dhat-\Dzero=\K\boldsymbol r_D$ shows that $\|\K\boldsymbol r_D\|_\infty$ bounds the discrepancy between the computed map and the exact center map. It is added directly to $\varepsilon_D$, without an additional $(1-\mu)^{-1}$ factor. Re-centering the perturbation identity at $\Dhat$ gives an alternative error decomposition.

The local map anchor $D\widetilde{\boldsymbol\psi}_{\mathrm{anc}}$ is the fixed center map used for comparison with $\Dstar$. Its error allowance $\varepsilon_{D,\mathrm{anc}}$ equals $\varepsilon_D$ when the anchor is $\Dzero$, and $\varepsilon_D+\|\K\boldsymbol r_D\|_\infty$ when the anchor is $\Dhat$. These local anchors form the global map anchor introduced in Subsection~\ref{subsec:numerical_conventions}. The approximate-inverse construction in Appendix~\ref{app:numerical} provides an alternative total bound with its center-map correction already included; the two constructions are used separately.

With module indices restored, $\Delta(D\boldsymbol\psi_m)=(D\boldsymbol\psi_m)^*-(D\boldsymbol\psi_m)_{\mathrm{anc}}$ denotes the error relative to the corresponding anchor in physical coordinates. The module allowances $\varepsilon_{p,m}$ and $\varepsilon_{D,\mathrm{anc},m}$ are converted using the scale matrices of Subsection~\ref{subsec:numerical_conventions}. This gives a passive-coordinate allowance vector $\boldsymbol b_{p,m}$ and a differential-map allowance matrix $\boldsymbol B_{D,m}$:
\begin{equation}
 |\boldsymbol p_m^*-\boldsymbol p_{m,0}|\leq\boldsymbol b_{p,m}:=
 \boldsymbol\Lambda_{m,p}\boldsymbol1\varepsilon_{p,m},\qquad |\Delta(D\boldsymbol\psi_m)|\leq\boldsymbol B_{D,m}:=
 \varepsilon_{D,\mathrm{anc},m}\boldsymbol\Lambda_{m,p}\boldsymbol1\boldsymbol1^{\mathsf T}\boldsymbol\Lambda_{m,a}^{-1}.
 \label{eq:physical_differential_allowance}
\end{equation}
Here $\boldsymbol p_{m,0}$ is the center in physical coordinates, and each $\boldsymbol1$ is an all-ones vector with the passive or active module dimension required by the adjacent scale matrix. The first inequality bounds each passive-coordinate error, while the second bounds each entry of the local differential-map error. The dense outer product gives a safe componentwise conversion of the induced infinity-norm allowance; additional local zeros can be retained where an analytical support argument is available.

\section{Structure-Preserving Path and Task Propagation}
\label{sec:task}
\subsection{Module support and task conventions}
\label{subsec:module_task_conventions}
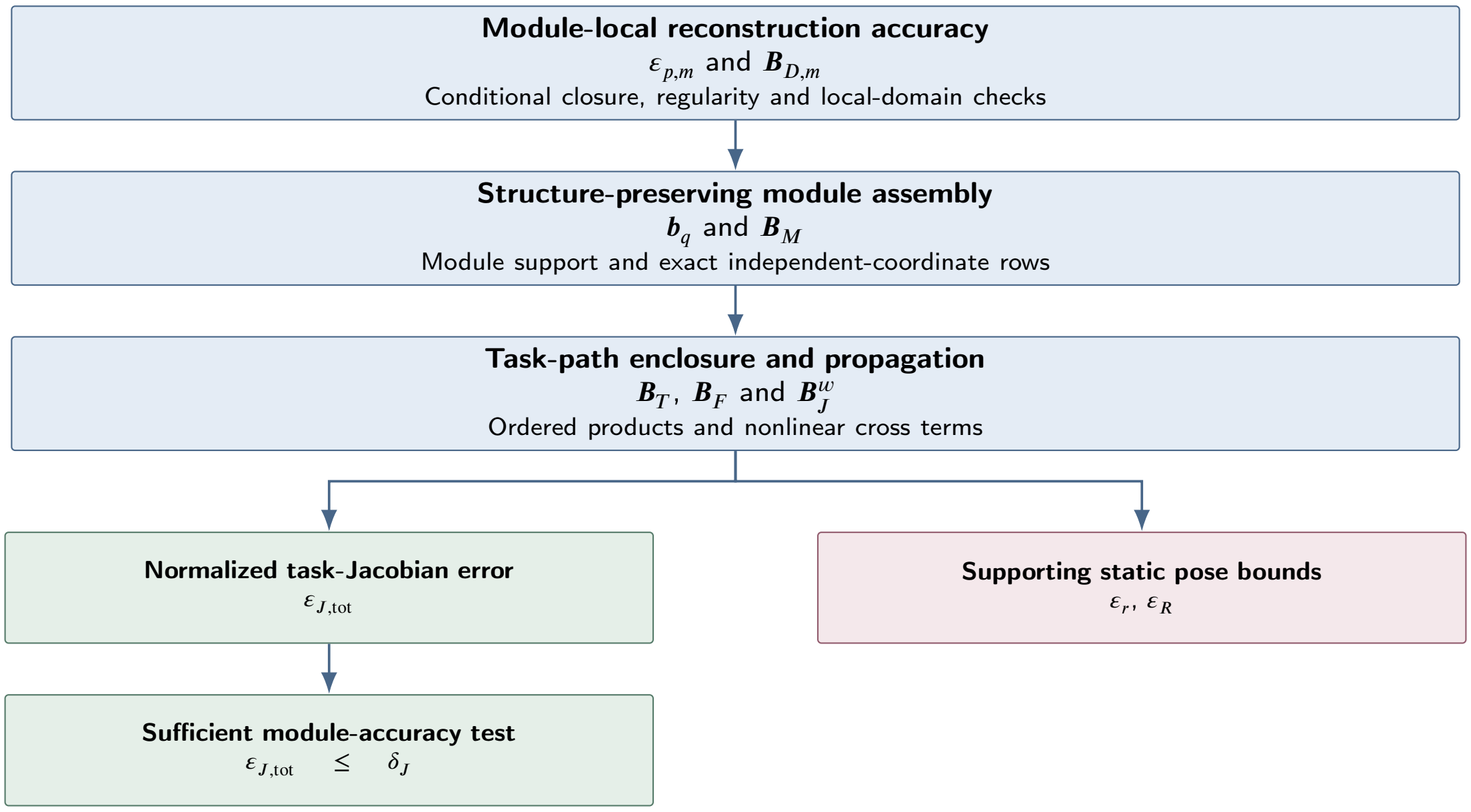
\begin{figure}[pos=!tb,abovecap=3pt,belowcap=1pt]
\centering
\definecolor{paStageBlue}{RGB}{228,236,245}
\definecolor{paStageSage}{RGB}{229,239,232}
\definecolor{paStageRose}{RGB}{244,232,235}
\definecolor{paLineBlue}{RGB}{72,102,136}
\definecolor{paLineSage}{RGB}{87,124,107}
\definecolor{paLineRose}{RGB}{146,92,110}
\clipbox{0cm 0cm 0cm 0cm}{%
\begin{tikzpicture}[
  >=Latex,
  every node/.style={font=\footnotesize,align=center,text=black},
  stage/.style={draw=paLineBlue,fill=paStageBlue,
                rounded corners=1pt,inner sep=3pt,
                minimum height=9mm,text width=.86\columnwidth},
  branch/.style={draw=paLineSage,fill=paStageSage,
                 rounded corners=1pt,inner sep=2.5pt,
                 minimum height=11mm,text width=.38\columnwidth,
                 font=\scriptsize},
  link/.style={-Latex,draw=paLineBlue,line width=.7pt}
]
\node[stage] (local) {\textbf{Module-local reconstruction accuracy}\\
  $\varepsilon_{p,m}$ and $\boldsymbol B_{D,m}$\\
  {\scriptsize Conditional closure, regularity and local-domain checks}};
\node[stage,below=5mm of local] (assembly)
  {\textbf{Structure-preserving module assembly}\\
   $\boldsymbol b_q$ and $\boldsymbol B_M$\\
   {\scriptsize Module support and exact independent-coordinate rows}};
\node[stage,below=5mm of assembly] (path)
  {\textbf{Task-path enclosure and propagation}\\
   $\boldsymbol B_T$, $\boldsymbol B_F$ and $\boldsymbol B_J^w$\\
   {\scriptsize Ordered products and nonlinear cross terms}};
\node[branch,below=8mm of path,xshift=-.245\columnwidth] (jac)
  {\textbf{Normalized task-Jacobian error}\\
   $\varepsilon_{J,\mathrm{tot}}$};
\node[branch,draw=paLineRose,fill=paStageRose,below=8mm of path,xshift=.245\columnwidth] (pose)
  {\textbf{Supporting static pose bounds}\\
   $\varepsilon_r$, $\varepsilon_R$};
\node[branch,below=5mm of jac] (budget)
  {\textbf{Sufficient module-accuracy test}\\
   $\varepsilon_{J,\mathrm{tot}}\leq\delta_J$};
\draw[link] (local) -- (assembly);
\draw[link] (assembly) -- (path);
\draw[link] (path.south) -- ++(0,-3mm) -| (jac.north);
\draw[link] (path.south) -- ++(0,-3mm) -| (pose.north);
\draw[link] (jac) -- (budget);
\end{tikzpicture}%
}%
\caption{Structure of the kinematic accuracy analysis. Conditional
module-local reconstruction and differential-map error allowances are
assembled while preserving coordinate support and propagated through the
task path to obtain a normalized reduced task-Jacobian error bound. This
bound provides sufficient module-specific accuracy requirements; static
end-point position and orientation allowances follow separately from the
task-path enclosure.}
\label{fig:accuracy_workflow}
\end{figure}

At a fixed independent configuration, the reconstructed passive centers form the global center $\boldsymbol q_0$. The assembly in~\eqref{eq:global-reconstruction}, evaluated with the local anchors of Subsection~\ref{subsec:differential_map_sensitivity}, gives $\Manc$, retaining the exact active-coordinate identity rows. The local allowances $\boldsymbol b_{p,m}$ and $\boldsymbol B_{D,m}$ then give the global coordinate-error vector $\boldsymbol b_q\in\R^{n_q}$ and map-error matrix $\boldsymbol B_M\in\R^{n_q\times n_a}$:
\begin{equation}
 \boldsymbol b_q=\sum_m\boldsymbol E_{p,m}\boldsymbol b_{p,m},\qquad
 \boldsymbol B_M=\sum_m\boldsymbol E_{p,m}\boldsymbol B_{D,m}\boldsymbol L_m.
 \label{eq:module_allowance_assembly}
\end{equation}
These nonnegative arrays satisfy $|\boldsymbol q^*-\boldsymbol q_0|\leq\boldsymbol b_q$ and $|\boldsymbol M^*-\Manc|\leq\boldsymbol B_M$. The fixed insertion matrices $\boldsymbol E_{p,m}$ and selectors $\boldsymbol L_m$ preserve each module's coordinate support, so the active-coordinate radii and corresponding map-row allowances are exactly zero. In compatible dimensionless coordinates, the module infinity-norm bounds imply $\|\boldsymbol q^*-\boldsymbol q_0\|_2\leq(\sum_m n_{m,p}\varepsilon_{p,m}^2)^{1/2}$, where $n_{m,p}$ is the number of passive coordinates in module $m$. Physical mixed-unit comparisons instead use $\boldsymbol b_q$ with the declared scales. Figure~\ref{fig:accuracy_workflow} summarizes the subsequent propagation.

The \emph{end point} is the point of interest fixed to the terminal body. An attached frame has its origin at this point, and its pose relative to the world frame is $\boldsymbol T=[\rot\ \boldsymbol r;\boldsymbol0^{\mathsf T}\ 1]$, with position $\boldsymbol r\in\R^3$ and orientation $\rot\in\SO(3)$. The spatial columns $\boldsymbol\eta_j=(\partial_j\boldsymbol T\boldsymbol T^{-1})^\vee$, where $\partial_j$ denotes differentiation with respect to $q_j$, form $\boldsymbol\eta=[\boldsymbol\eta_1\ \cdots\ \boldsymbol\eta_{n_q}]=\col(\boldsymbol\Omega,\boldsymbol V)$. The matrices $\boldsymbol\Omega,\boldsymbol V\in\R^{3\times n_q}$ are its angular and linear spatial-twist blocks. The skew-symmetric matrix $[\boldsymbol r]_\times$ is defined by $[\boldsymbol r]_\times\boldsymbol x=\boldsymbol r\times\boldsymbol x$. The standard relation between spatial twists and point velocities~\cite{lynch2017modern}, together with the coordinate reduction in~\eqref{eq:global-reconstruction}, yields the translation-first end-point velocity map and reduced task Jacobians
\begin{equation}
 \boldsymbol F=\begin{bmatrix}\boldsymbol V-[\boldsymbol r]_\times\boldsymbol\Omega\\\boldsymbol\Omega\end{bmatrix},
 \qquad\boldsymbol J^w=\boldsymbol F\boldsymbol M,
 \qquad\boldsymbol J=\boldsymbol Q\boldsymbol J^w.
 \label{eq:task_map_factorization}
\end{equation}
Here $\boldsymbol F\in\R^{6\times n_q}$ maps full-coordinate rates to the linear velocity of the end point followed by the angular velocity of its attached frame, both expressed in world coordinates. The subtraction $[\boldsymbol r]_\times\boldsymbol\Omega$ accounts for the offset between the end point and the world origin. The factor $\boldsymbol M$ is the active-to-full-coordinate map; thus $\boldsymbol J^w\in\R^{6\times n_a}$ maps independent-coordinate rates to these world-frame velocities. The orthogonal matrix $\boldsymbol Q$ specifies the task basis: $\boldsymbol Q=\boldsymbol I_6$ retains world components, whereas $\boldsymbol Q=\blkdiag(\boldsymbol I_3,\rot_{\mathrm{meas}}^{\mathsf T})$ expresses angular components in a measured end-point frame with orientation $\rot_{\mathrm{meas}}\in\SO(3)$. The same $\boldsymbol Q$ is held fixed for both maps being compared; independently recomputed bases require an additional perturbation allowance.

The translational and angular task scales are denoted by $\chi_r>0$ in metres and $\chi_R>0$ in radians. Together with a fixed positive diagonal independent-coordinate scale $\boldsymbol H\in\R^{n_a\times n_a}$, they define the task normalization matrix $\SX$ and the normalized Jacobian:
\begin{equation}
 \SX=\blkdiag(\chi_r^{-1}\boldsymbol I_3,\chi_R^{-1}\boldsymbol I_3),
 \qquad\widetilde{\boldsymbol J}=\SX\boldsymbol J\boldsymbol H.
 \label{eq:task_normalization}
\end{equation}
For an independent-coordinate velocity $\boldsymbol u\in\R^{n_a}$ and a translation-first task velocity $\boldsymbol v\in\R^6$, the scaled rates $\boldsymbol H^{-1}\boldsymbol u$ and $\SX\boldsymbol v$ have units $\mathrm{s}^{-1}$, making $\widetilde{\boldsymbol J}$ dimensionless. A positive diagonal weighting matrix $\boldsymbol G$ for independent-coordinate velocities gives the admissible choice $\boldsymbol H=\boldsymbol G^{-1/2}$. The isotropic scaling within each task block makes $\SX$ commute with $\boldsymbol Q$, which preserves the Euclidean norm. The scales are fixed before comparison as in Subsection~\ref{subsec:numerical_conventions}; a task-position-box diagonal and a prescribed rotation excursion provide possible choices for $\chi_r$ and $\chi_R$, respectively.

\subsection{Task-Jacobian bounds and comparison}
\label{subsec:task_jacobian_bounds}
Propagation of $\boldsymbol b_q$ through the ordered transformation path gives the entrywise allowance $|\boldsymbol F^*-\boldsymbol F_0|\leq\boldsymbol B_F$, where $\boldsymbol F_0=\boldsymbol F(\boldsymbol q_0)$ and $\boldsymbol F^*=\boldsymbol F(\boldsymbol q^*)$. The construction in Appendix~\ref{app:task_path_bounds} combines transform and derivative allowances using a finite-product bound. For conformable factors $\boldsymbol X,\boldsymbol Y$ with center values $\boldsymbol X_0,\boldsymbol Y_0$ and nonnegative bounds $|\boldsymbol X-\boldsymbol X_0|\leq\boldsymbol B_X$, $|\boldsymbol Y-\boldsymbol Y_0|\leq\boldsymbol B_Y$, the product error is bounded by
\begin{equation}
 \boldsymbol B_{XY}=|\boldsymbol X_0|\boldsymbol B_Y+
 \boldsymbol B_X|\boldsymbol Y_0|+\boldsymbol B_X\boldsymbol B_Y.
 \label{eq:componentwise_product_allowance}
\end{equation}
Thus $|\boldsymbol X\boldsymbol Y-\boldsymbol X_0\boldsymbol Y_0|\leq\boldsymbol B_{XY}$. The final term retains the product of the two finite perturbations. Repeated coordinate occurrences and shared inputs are accommodated by the box enclosure, with conservatism arising when their variations are treated independently.

\begin{theorem}[Reconstruction-to-task allowance]
\label{thm:task_allowance}
For nonnegative allowances $\boldsymbol B_F$ and $\boldsymbol B_M$ satisfying $|\boldsymbol F^*-\boldsymbol F_0|\leq\boldsymbol B_F$ and $|\boldsymbol M^*-\Manc|\leq\boldsymbol B_M$ at the same active coordinates, on the stated branch and in the stated units, the task-map anchors are $\Janc=\boldsymbol Q\boldsymbol F_0\Manc$ and $\boldsymbol J^{w}_{\mathrm{anc}}=\boldsymbol F_0\Manc$. With the fixed basis and normalization of Subsection~\ref{subsec:module_task_conventions}, the entrywise world-frame allowance $\boldsymbol B_J^w\in\R^{6\times n_a}$ and scalar normalized allowance $\varepsilon_J$ satisfy
\begin{equation}
 \boldsymbol B_J^w=\boldsymbol B_F|\Manc|
       +|\boldsymbol F_0|\boldsymbol B_M+\boldsymbol B_F\boldsymbol B_M,\quad |\boldsymbol J^{w,*}-\boldsymbol J^{w}_{\mathrm{anc}}|\leq\boldsymbol B_J^w,\quad \|\widetilde{\boldsymbol J}^*-\widetilde{\boldsymbol J}_{\mathrm{anc}}\|_2\leq
 \varepsilon_J:=\|\SX\boldsymbol B_J^w\boldsymbol H\|_2.
 \label{eq:task_jacobian_allowance}
\end{equation}
\end{theorem}
\begin{proof}
The perturbations $\Delta\boldsymbol F=\boldsymbol F^*-\boldsymbol F_0$ and $\Delta\boldsymbol M=\boldsymbol M^*-\Manc$ give the product difference $(\boldsymbol F_0+\Delta\boldsymbol F)(\Manc+\Delta\boldsymbol M)-\boldsymbol F_0\Manc$. Its expansion and componentwise absolute values yield the stated $\boldsymbol B_J^w$, including the cross term. For any error matrix $\boldsymbol E$ with $|\boldsymbol E|\leq\boldsymbol B$ and $\boldsymbol B\geq0$, the inequality $|\boldsymbol E\boldsymbol x|\leq\boldsymbol B|\boldsymbol x|$ holds for every compatible vector $\boldsymbol x$; taking Euclidean norms gives $\|\boldsymbol E\|_2\leq\|\boldsymbol B\|_2$. Application to the positively scaled allowance, together with $\SX\boldsymbol Q=\boldsymbol Q\SX$ and orthogonality of $\boldsymbol Q$, yields the normalized bound. The products retain the zeros imposed by module support and coordinates absent from the end-point path.
\end{proof}

The stored computed task Jacobian $\widehat{\boldsymbol J}$ may also differ from the anchor $\Janc$. This remaining discrepancy is $\boldsymbol C_J$, and its normalized contribution gives the total allowance $\varepsilon_{J,\mathrm{tot}}$ relative to $\widehat{\widetilde{\boldsymbol J}}=\SX\widehat{\boldsymbol J}\boldsymbol H$:
\begin{equation}
 \boldsymbol C_J=\widehat{\boldsymbol J}-\Janc,\qquad \|\widetilde{\boldsymbol J}^*-\widehat{\widetilde{\boldsymbol J}}\|_2
 \leq\varepsilon_{J,\mathrm{tot}}:=\varepsilon_J+
 \|\SX\boldsymbol C_J\boldsymbol H\|_2.
 \label{eq:total_task_map_discrepancy}
\end{equation}
The anchor choice determines this error accounting. In the numerical comparisons, $\Manc$ contains the stored local maps $\widehat{D\boldsymbol\psi}_m$, whose center-map discrepancies are already included in $\boldsymbol B_M$ through Section~\ref{sec:local}. Accordingly, $\boldsymbol C_J$ accounts only for separately supplied path, basis, or assembly evaluation errors. Taking this remaining contribution as zero follows the nominal-evaluation convention of Subsection~\ref{subsec:numerical_conventions}. With exact local center maps as anchors, $\boldsymbol B_M$ instead encloses reconstruction error, and the computation error is accounted for separately.

A scalar comparator is obtained from the same componentwise enclosures by reducing the two factors to compatible spectral-norm bounds. A fixed positive diagonal full-coordinate scale $\boldsymbol\Lambda_q\in\R^{n_q\times n_q}$ defines the nonnegative normalized anchor factors $\overline{\boldsymbol F}_{0,\mathrm{abs}}=\SX|\boldsymbol F_0|\boldsymbol\Lambda_q$ and $\overline{\boldsymbol M}_{\mathrm{anc},\mathrm{abs}}=\boldsymbol\Lambda_q^{-1}|\Manc|\boldsymbol H$. The subscript $\mathrm{abs}$ indicates that entrywise absolute values are taken before the spectral norms. The corresponding scalar variation allowances are $\delta_F=\|\SX\boldsymbol B_F\boldsymbol\Lambda_q\|_2$ and $\delta_M=\|\boldsymbol\Lambda_q^{-1}\boldsymbol B_M\boldsymbol H\|_2$. They give the dimensionless normwise task-Jacobian allowance
\begin{equation}
 \varepsilon_J^{\mathrm{nw}}=
 \delta_F\|\overline{\boldsymbol M}_{\mathrm{anc},\mathrm{abs}}\|_2+
 \|\overline{\boldsymbol F}_{0,\mathrm{abs}}\|_2\delta_M+\delta_F\delta_M.
 \label{eq:normwise_comparator}
\end{equation}
The estimate follows from the triangle inequality and induced-norm submultiplicativity applied to the scaled componentwise product bound. With identical $\boldsymbol B_F$, $\boldsymbol B_M$, and anchor factors, these inequalities give $\varepsilon_J\leq\varepsilon_J^{\mathrm{nw}}$. Entrywise absolute anchor factors are essential to this comparison, because taking norms of signed center matrices would retain cancellations already removed in the componentwise enclosure. Subsection~\ref{subsec:comparison_protocol} uses this shared-input comparison to assess the effect of preserving structure.

\subsection{Supporting task-kinematic consequences}
\label{subsec:task_kinematic_consequences}
\begin{corollary}[Specified-velocity accuracy]
\label{cor:task_rate}
For a fixed finite independent-coordinate velocity $\boldsymbol u$ and a prescribed task rate $\boldsymbol v$ in the same basis, the directional allowance $\beta(\boldsymbol u)$ bounds the normalized same-input map discrepancy and gives
\begin{equation}
 \beta(\boldsymbol u)=\|\SX\boldsymbol B_J^w|\boldsymbol u|\|_2+
                         \|\SX\boldsymbol C_J\boldsymbol u\|_2,\qquad \|\SX(\boldsymbol J^*\boldsymbol u-\boldsymbol v)\|_2
 \leq\beta(\boldsymbol u)+\|\SX(\widehat{\boldsymbol J}\boldsymbol u-\boldsymbol v)\|_2.
 \label{eq:specified_velocity_accuracy}
\end{equation}
The result applies to any supplied $\boldsymbol u$, independently of task rank or the procedure used to select it.
\end{corollary}
\begin{proof}
The difference $\boldsymbol J^*\boldsymbol u-\boldsymbol v$ separates into the same-input map discrepancy $(\boldsymbol J^*-\widehat{\boldsymbol J})\boldsymbol u$ and the computed target-rate residual $\widehat{\boldsymbol J}\boldsymbol u-\boldsymbol v$. The componentwise bound in Theorem~\ref{thm:task_allowance}, the definition of $\boldsymbol C_J$, and preservation of the scaled Euclidean norm under the fixed basis change bound the first term by $\beta(\boldsymbol u)$. The triangle inequality then gives the stated result.
\end{proof}
The allowance $\beta(\boldsymbol u)$ in~\eqref{eq:specified_velocity_accuracy} concerns reconstruction error at the same supplied input, while $\|\SX(\widehat{\boldsymbol J}\boldsymbol u-\boldsymbol v)\|_2$ retains the computed discrepancy from the target rate, including any residual of a numerically computed null-space direction. The translational and angular parts of the same-input discrepancy have norm allowances $10^3\chi_r\beta(\boldsymbol u)$ mm/s and $\chi_R\beta(\boldsymbol u)$ rad/s, respectively. For a prescribed nonnegative input-rate bound $\alpha$ satisfying $\|\boldsymbol H^{-1}\boldsymbol u\|_2\leq\alpha$, the operator estimate gives $\|\SX(\boldsymbol J^*-\widehat{\boldsymbol J})\boldsymbol u\|_2\leq\varepsilon_{J,\mathrm{tot}}\alpha$. Here $\alpha$ and the normalized rate-error allowance have units $\mathrm{s}^{-1}$, and $\varepsilon_{J,\mathrm{tot}}\alpha$ can replace $\beta(\boldsymbol u)$ in the physical-unit expressions when only this input-rate bound is specified.

\begin{proposition}[Separate task-rank margin]
\label{prop:task_rank}
For a task with no more rows than independent coordinates, $\sigma_{\min}$ denotes the smallest singular value of its normalized Jacobian. The positive margin $\underline\sigma$ gives the sufficient full-row-rank condition
\begin{equation}
 \underline\sigma=\sigma_{\min}(\widetilde{\boldsymbol J}_{\mathrm{anc}})-\varepsilon_J>0
 \quad\Longrightarrow\quad
 \sigma_{\min}(\widetilde{\boldsymbol J}^*)\geq\underline\sigma>0.
 \label{eq:task_rank_margin}
\end{equation}
\end{proposition}
\begin{proof}
The singular-value perturbation inequality~\cite{stewart1990svd} bounds the change in the smallest singular value by $\|\widetilde{\boldsymbol J}^*-\widetilde{\boldsymbol J}_{\mathrm{anc}}\|_2$, which is at most $\varepsilon_J$ by Theorem~\ref{thm:task_allowance}. Transposition gives the same bound for a wide matrix.
\end{proof}
This margin assesses task rank separately from passive-closure regularity. A nonpositive margin leaves task rank undetermined by this sufficient test. A corresponding test at $\widehat{\widetilde{\boldsymbol J}}$ uses the total allowance $\varepsilon_{J,\mathrm{tot}}$.

\begin{corollary}[Static end-point pose accuracy]
\label{cor:static_pose}
At the same fixed active configuration, the end-point transform enclosure $|\boldsymbol T^*-\boldsymbol T_0|\leq\boldsymbol B_T$ is obtained from Appendix~\ref{app:task_path_bounds}, with $\boldsymbol T_0=\boldsymbol T(\boldsymbol q_0)$ and $\boldsymbol T^*=\boldsymbol T(\boldsymbol q^*)$. Its translation-column allowance $\boldsymbol b_r=\boldsymbol B_T(1{:}3,4)\in\R^3$ and rotation-block allowance $\boldsymbol B_{\mathcal R}=\boldsymbol B_T(1{:}3,1{:}3)\in\R^{3\times3}$ give position and orientation allowances $\varepsilon_r$ and $\varepsilon_R$. Here $d_{\SO(3)}$ is the shortest relative rotation angle in $[0,\pi]$, and $\|\cdot\|_F$ denotes the Frobenius norm. The bounds are
\begin{equation}
 \|\boldsymbol r^*-\boldsymbol r_0\|_2\leq\varepsilon_r:=\|\boldsymbol b_r\|_2,\qquad d_{\SO(3)}(\rot^*,\rot_0)\leq\varepsilon_R:=2\arcsin\!\left(\min\!\left\{1,\frac{\|\boldsymbol B_{\mathcal R}\|_F}{2\sqrt2}\right\}\right).
 \label{eq:static_pose_allowance}
\end{equation}
\end{corollary}
\begin{proof}
The translation-column enclosure yields the Euclidean position bound. For $\rot^*,\rot_0\in\SO(3)$, the identity $\|\rot^*-\rot_0\|_F=2\sqrt2\sin(d_{\SO(3)}(\rot^*,\rot_0)/2)$ relates the matrix difference to the relative rotation angle~\cite{huynh2009metrics}. The rotation-block enclosure and monotonicity of $\arcsin$ on $[0,1]$ then give the angular bound, with saturation at $\pi$ when the matrix allowance reaches its full angular range.
\end{proof}
For a relative angle below $\pi$, $d_{\SO(3)}$ equals the norm of the principal rotation vector $\Log(\rot_0^{\mathsf T}\rot^*)^\vee$. Thus~\eqref{eq:static_pose_allowance} converts a matrix enclosure into an angle, with end-point orientation referring to the attached frame. The physical allowances are $10^3\varepsilon_r$ mm for position and $\varepsilon_R$ rad for orientation; their normalized counterparts are $\varepsilon_r/\chi_r$ and $\varepsilon_R/\chi_R$. They isolate numerical reconstruction error at fixed active coordinates, with physical-model and execution errors remaining separate contributors to positioning accuracy. Use of a computed end-point pose additionally requires a center-evaluation allowance and valid $\SO(3)$ rotation matrices.

\section{Task-Accuracy Budgets and Numerical Assessment}
\label{sec:feasibility}
\subsection{Accuracy requirements}
\label{subsec:accuracy_requirements}
A prescribed accuracy requirement for the normalized reduced task Jacobian is expressed by a dimensionless budget $\delta_J>0$, with the task and independent-coordinate scales held fixed. The module-local allowances are sufficient for this requirement when their propagated total bound satisfies
\begin{equation}
 \varepsilon_{J,\mathrm{tot}}\leq\delta_J.
 \label{eq:jacobian_budget_test}
\end{equation}
The operator bound controls the map discrepancy for every independent-coordinate velocity through the same-input estimates of Subsection~\ref{subsec:task_kinematic_consequences}. Supporting static-pose requirements can be specified by a position budget $\delta_{r,\mathrm{mm}}$ in millimetres and an orientation budget $\delta_R$ in radians. Corollary~\ref{cor:static_pose} gives the corresponding sufficient checks $10^3\varepsilon_r\leq\delta_{r,\mathrm{mm}}$ and $\varepsilon_R\leq\delta_R$. Thus $10^3\varepsilon_r<1$ and $\varepsilon_R<10^{-3}$ would satisfy illustrative numerical reconstruction-error requirements below 1 mm and 0.001 rad, respectively. These examples are possible specifications, not application requirements prescribed for the evaluated robot. For a supplied independent-coordinate velocity $\boldsymbol u$ and target task rate $\boldsymbol v$, a normalized rate-error budget $\delta_v>0$, in $\mathrm{s}^{-1}$, is met when
\begin{equation}
 \beta(\boldsymbol u)+\|\SX(\widehat{\boldsymbol J}\boldsymbol u-\boldsymbol v)\|_2\leq\delta_v.
 \label{eq:task_budget_test}
\end{equation}
Here the directional allowance $\beta(\boldsymbol u)$ is defined in~\eqref{eq:specified_velocity_accuracy}, and the second term accounts for the computed discrepancy from the target rate. The pose and rate checks support the primary task-Jacobian requirement~\eqref{eq:jacobian_budget_test}.

A \emph{feasible module-accuracy vector} consists of the passive-coordinate and differential-map allowances $\varepsilon_{p,m}$ and $\varepsilon_{D,\mathrm{anc},m}$ for the individual modules, satisfying their local validity conditions and the propagated task-Jacobian budget. This defines a sufficient set from which an allocation may be selected. Relating an allocation to solver tolerances requires a justified exact-residual budget and reevaluation at each reconstruction center. A bound is reported as \emph{unavailable} when its local sufficient conditions cannot be established; this denotes an inconclusive accuracy test, separately from the solver's closure-acceptance status.

\subsection{Comparison and reference protocol}
\label{subsec:comparison_protocol}
Each paired comparison uses a common branch, selected closure rows, center convention, task path, units, normalization, and numerical-reference protocol. Task components are expressed in the world frame in translation-first order. The structured allowance~\eqref{eq:task_jacobian_allowance} and normwise estimate~\eqref{eq:normwise_comparator} use identical componentwise enclosures, so their comparison isolates the effect of retaining structure on the bound magnitude. The normwise expression reuses those inputs and is not a separately implemented or timed end-to-end estimator. Reconstruction and bound evaluation are timed separately, excluding numerical-reference refinement; each reported time is the median of three repetitions.

A numerical reference obtained by further correction retains its own uncertainty. If $d$ is the distance between a reconstructed quantity and its numerical reference, and $\eta_{\mathit{ref}}$ bounds the reference error in the same norm, the exact discrepancy lies in $[\max(0,d-\eta_{\mathit{ref}}),d+\eta_{\mathit{ref}}]$. The present evaluations follow the nominal-double convention of Subsection~\ref{subsec:numerical_conventions}. Trial comparisons use references refined with a maximum of 12 Newton iterations and fixed selected closure rows. The reference correction tests step factors $1$, $1/2$, $1/4$, and $1/8$, retaining strictly interior, residual-decreasing steps and the original full-closure gate. Repeats allow up to 24 iterations and check agreement within $10^{-9}$ in the scaled coordinate infinity norm and normalized Jacobian spectral norm. Conditional allowances evaluated at the reference centers provide an additional consistency check. The directional comparison in Subsection~\ref{subsec:directional_validation} uses the longer refinement for its center and perturbed configurations. These calculations assess numerical consistency, with the exact-reference error floor remaining unresolved.

\subsection{Configurations and numerical setup}
\label{subsec:numerical_setup}
\begin{figure}[pos=!tb,abovecap=3pt,belowcap=1pt]
\centering
\includegraphics[width=0.90\linewidth,trim={0.6cm 0.1cm 0.6cm 0.5cm},clip]{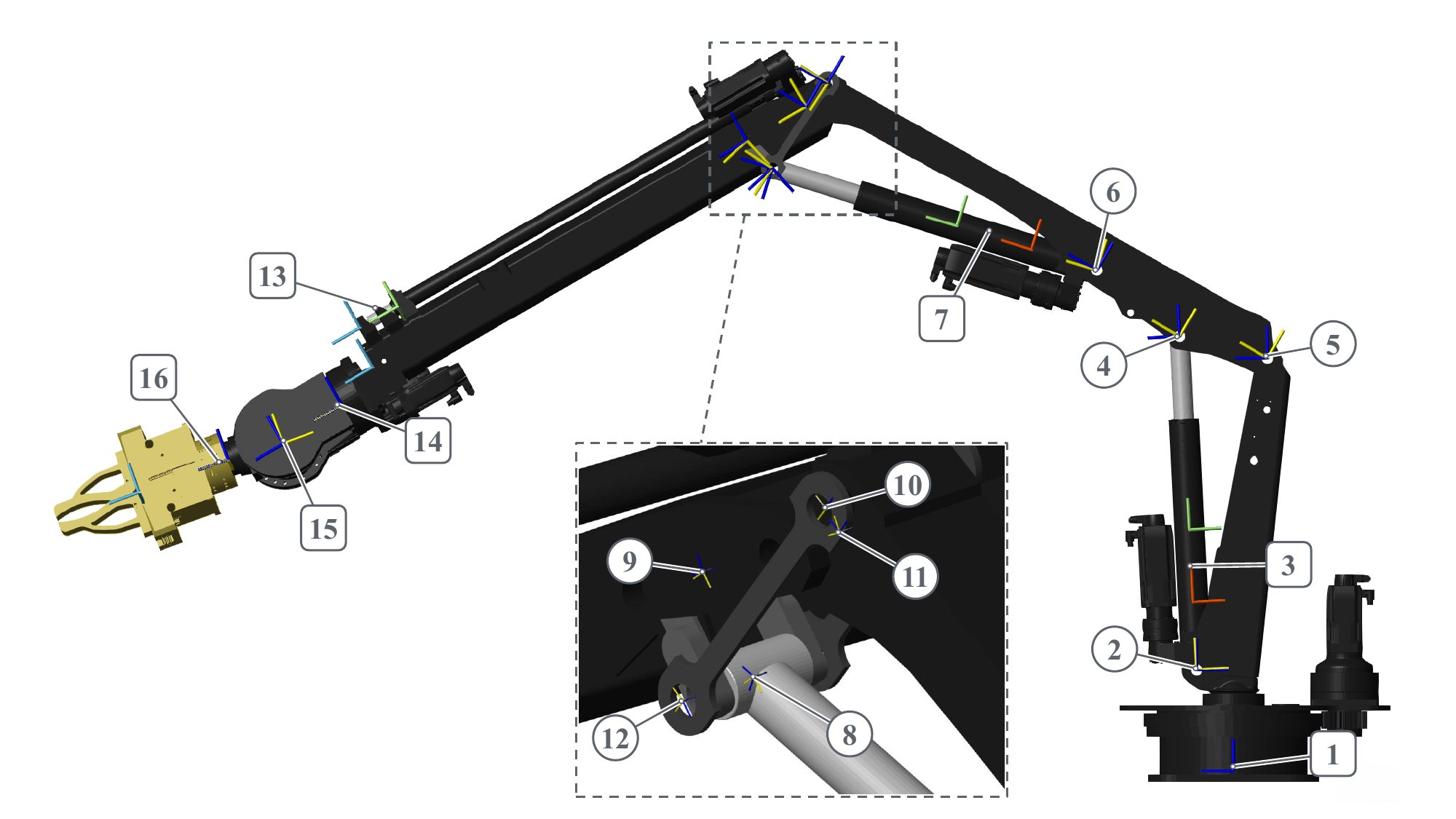}
\caption{Simscape Multibody model of the HIAB-046 seven-degree-of-freedom manipulator. Label $i$ identifies the joint associated with coordinate $q_i$; square labels denote independent coordinates, and circular labels denote passive coordinates. Dashed boxes and their connecting line identify the enlarged four-bar linkage detail. Physical parameters are omitted under a non-disclosure agreement.}
\label{fig:simscape-robot-model}
\end{figure}

\begin{figure}[pos=!tb,abovecap=3pt,belowcap=1pt]
\centering
\includegraphics[width=0.90\linewidth,trim={0cm 4.6cm 0cm 2.2cm},clip]{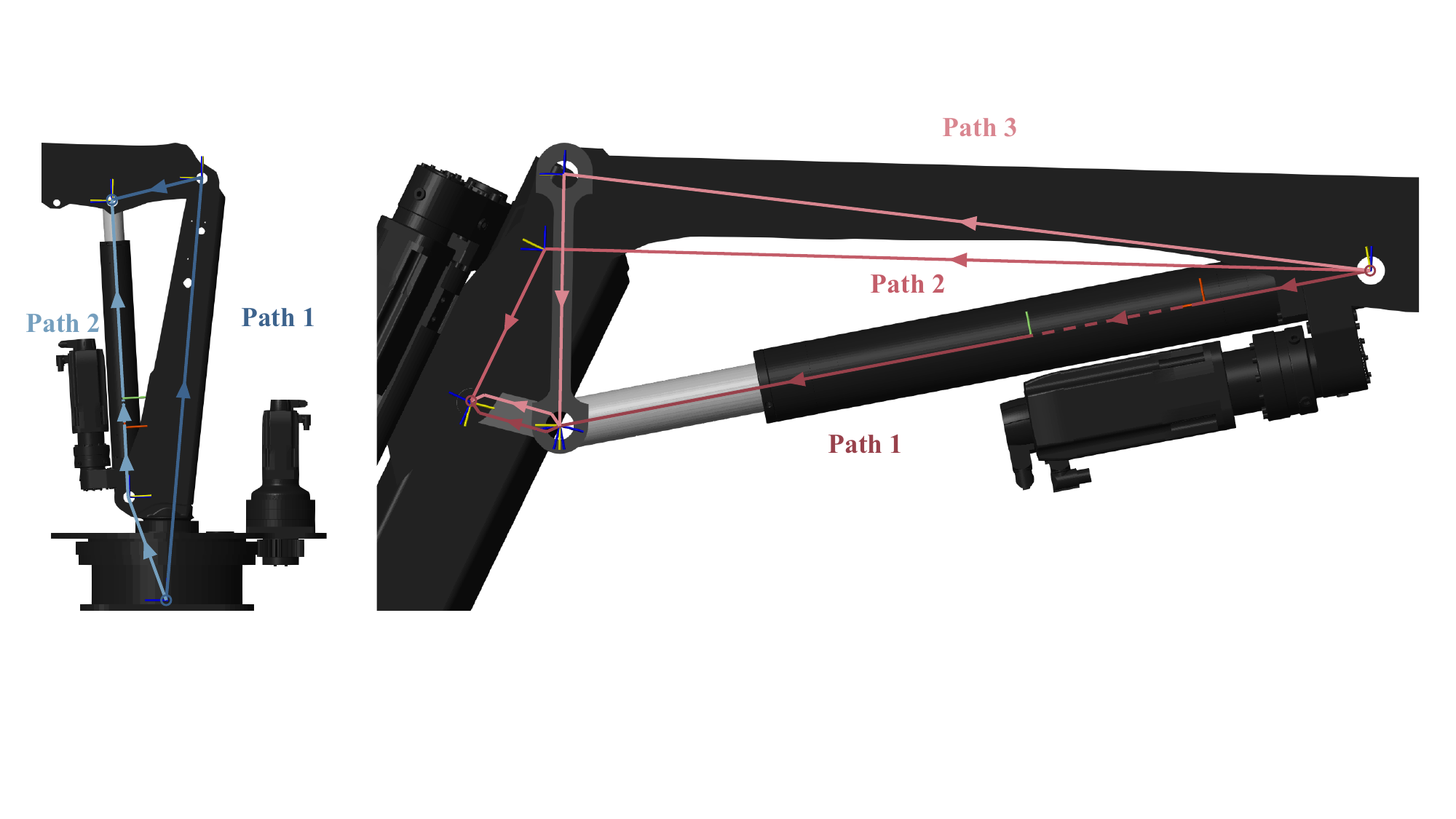}
\caption{Physical views and ordered kinematic paths of the closed-chain modules: Module 1 (\textit{left}), with two paths in blue shades, and Module 2 (\textit{right}), with three paths in red shades. Arrowheads indicate path traversal, not physical motion or velocity. Dashed segments mark the base-to-follower transformations of prismatic joints. Separately drawn schematic arrows on shared portions distinguish paths without representing additional links.}
\label{fig:closed-chain-modules}
\end{figure}

\begin{figure}[pos=!tb,abovecap=3pt,belowcap=1pt]
\centering
\includegraphics[width=0.95\linewidth,trim={0cm 4.3cm 0cm 4.75cm},clip]{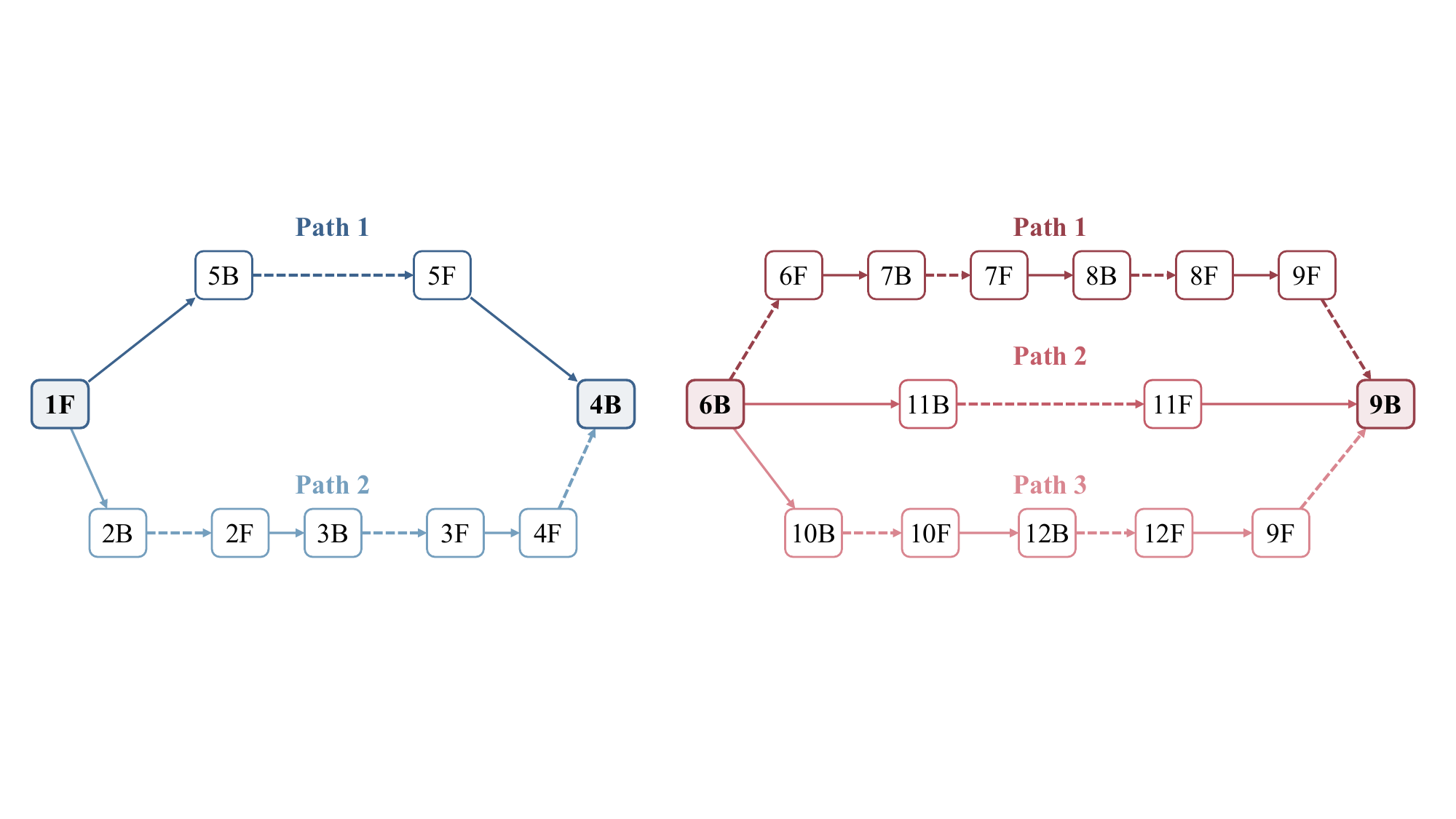}
\caption{Local-transformation graphs corresponding to Fig.~\ref{fig:closed-chain-modules}: Module 1 (\textit{left}) and Module 2 (\textit{right}). Path colors and numbering match the physical views. Nodes denote frames, with $i\mathrm{B}$ and $i\mathrm{F}$ identifying the base and follower frames of joint $i$. Solid arrows denote rigid transformations within links; dashed arrows denote joint-motion transformations of revolute or prismatic joints. Bold labels and shaded nodes mark the common endpoint frames. The two occurrences of $9\mathrm{F}$ in Module 2 represent the same physical frame.}
\label{fig:path-transformation-graphs}
\end{figure}

The assessment uses nine fixed active-coordinate targets of the HIAB-046 7-DoF heavy-duty robotic manipulator considered in the path assembly kinematic reduction study~\cite{dastranj2026}. Figure~\ref{fig:simscape-robot-model} identifies the joint coordinates and the enlarged four-bar detail; Fig.~\ref{fig:closed-chain-modules} shows the ordered paths of the two physical closed-chain modules, and Fig.~\ref{fig:path-transformation-graphs} shows their corresponding frame-transformation graphs. The independent coordinates are $(q_1,q_3,q_7,q_{13},q_{14},q_{15},q_{16})$, while the passive sets are $(q_2,q_4,q_5)$ and $(q_6,q_8,q_9,q_{10},q_{11},q_{12})$ for Modules 1 and 2, respectively. The corresponding local active inputs are $q_3$ and $q_7$, so the local differential maps have dimensions $3\times1$ and $6\times1$. Coordinates $q_3$, $q_7$, and $q_{13}$ are translational (m), and the other coordinates are angular (rad).

All cases use one accepted predecessor and the same declared coordinate and task scales. The diagonal entries of $\boldsymbol\Lambda_q$ are the widths of the declared coordinate intervals in their native units; $\boldsymbol\Lambda_{m,a}$, $\boldsymbol\Lambda_{m,p}$, and $\boldsymbol H$ use the corresponding subsets. The task scales $\chi_r$ and $\chi_R$ are the Euclidean diagonal length of the declared task-position box (m) and the declared angular excursion (rad), respectively. For each closure element, the auxiliary length $L$ of Appendix~\ref{app:path_majorants} is the larger translation majorant of its two paths over the passive intervals at the fixed active coordinates. The selected angular and translational rows use diagonal scales $1$ (rad) and $L$ (m), respectively, in $\boldsymbol\Lambda_{m,c}$. These residual weights are fixed within each local analysis but may vary with the active configuration. Numerical magnitudes of the model-derived scales and interval limits are withheld with the protected model data. Independent numerical replication requires access to these declarations and the nominal model.

The predecessor has $q_1=0$ rad, $q_3=0.253400$ m, and $q_7=0.275600$ m; the remaining independent coordinates are $q_{13}=0.050000$ m, $q_{14}=q_{16}=0.17450000$ rad, and $q_{15}=1.74500000$ rad. The original configurations C01--C03 retain these values except for $(q_1,q_3)$, which are respectively $(0.06283185\ \mathrm{rad},0.253400\ \mathrm{m})$, $(0\ \mathrm{rad},0.243400\ \mathrm{m})$, and $(0.04398230\ \mathrm{rad},0.245900\ \mathrm{m})$.

The six additional configurations C04--C09 were specified before their outcomes were evaluated. One independent coordinate per module was selected from the largest column norm of its normalized passive differential map at the predecessor, giving $q_3$ and $q_7$. C04--C07 use the four sign pairs $(++),(+-),(-+),(--)$ at 2\% of the declared working-interval widths; C08 and C09 use $(++)$ and $(--)$ at 5\%. The resulting displacements from the predecessor are $\pm0.010$ m and $\pm0.025$ m in each selected coordinate, respectively, with all other independent coordinates unchanged. The planned steps were within the declared intervals, and all six configurations were retained. Both modules change passive configuration in these cases, whereas the Module 2 baseline coordinates are unchanged across C01--C03. Passive predictors from the common predecessor initialize direct physical-closure correction. Homotopy fallback is disabled for the additional cases and the directional checks; it was not used in any of the recorded trials.

The original numerical assessment and the subsequent diagnostic replay used MATLAB R2025b (64-bit Windows, release 25.2.0.2998904). The timings belong solely to the assessment of all nine cases; the replay evaluated the same cases once per trial without new timings or a repeated directional campaign. Both records identify an Intel64 processor, family 6, model 183, stepping 1; the replay additionally records an Intel Core i9-14900K. The exact processor model name was not recorded for the original timing run. The baseline corrector and selected-closure tolerances were $10^{-9}$, with full-closure acceptance fixed at $10^{-8}$. At each target, three global trials multiply the corrector tolerance by $f_g\in\{0.1,1,10\}$, starting from the common predecessor and retaining the closure-acceptance conditions. Their bounds use the residuals of the resulting reconstruction centers. Separately, the fixed-center budget assessment uses hypothetical selected-residual allowances $\tau_m=f_m\,10^{-9}$, with $f_1,f_2\in\{0.1,1,10\}$, at one baseline reconstruction per configuration. The complete assessment therefore contains 27 reconstruction trials and 81 module-budget combinations.

Direct correction allows at most eight iterations, with four line-search trials at successive halvings from a unit step. The trial passive-coordinate displacement must exceed $10^{-13}(1+\|\boldsymbol p_m\|_\infty)$ in the native coordinates. A Newton step uses the selected raw block when its reciprocal condition estimate exceeds $10^{-11}$; otherwise the damped normal-equation step uses damping $10^{-8}$. The relative rank threshold and selected-block reciprocal condition threshold are both $10^{-10}$. Stopping and closure gates act on the raw residuals, whereas the error bounds use their declared normalization. The observed-residual bound evaluator screens selected residuals at the full-closure threshold $10^{-8}$ and uses the actual center residual; the hypothetical-budget evaluator instead uses $\tau_m$ as its selected-residual gate and converts that budget with the selected row scales. Neither changes the solver acceptance criteria.

The diagnostic replay records the ordered physical selectors at all baseline, reference, and trial centers. Indices are one-based in the stacked angular-first six-component residuals of the closure elements. Module 1 uses $(3,4,5)$ throughout. Module 2 uses $(9,3,10,5,11,4)$ except at C08, where it uses $(9,3,11,5,4,10)$, a permutation of the same set. The bound evaluation uses the computed-center construction of Appendix~\ref{app:numerical}; the numerical work limits and safeguards are specified there. In the replay, all 27 trial scalar comparisons and all 81 budget-cell availability statuses agree with the original records, including all 45 available budget bounds, within the prescribed audit tolerances (relative $10^{-8}$ plus absolute $10^{-13}$ for bounds, relative $10^{-6}$ plus absolute $10^{-12}$ for observed Jacobian discrepancies).

Table~\ref{tab:host_three_case} separates the repeated original configurations from the additional cases. All nine baseline configurations and all 27 global trials were accepted, with available conditional task-Jacobian bounds. Each trial retained the selected rows of its own baseline reconstruction.

\begin{table}[pos=!tb,abovecap=4pt,belowcap=1pt]
\caption{Outcomes of the fixed-configuration assessment. C01--C03 repeat the original configurations; C04--C09 are the six additional configurations. Counts retain all planned cases, trials, and budget combinations. Timing ranges span the per-trial medians of three repetitions and exclude numerical-reference refinement.}
\label{tab:host_three_case}
\centering
\begingroup
\small
\setlength{\tabcolsep}{6pt}
\renewcommand{\arraystretch}{1.12}
\begin{tabular}{@{}p{0.57\linewidth}rr@{}}
\toprule
Quantity or comparison & C01--C03 & C04--C09 \\
\midrule
Accepted configurations & $3/3$ & $6/6$ \\
Accepted global trials; homotopy fallbacks & $9/9$; $0$ & $18/18$; $0$ \\
Available task-Jacobian trial bounds & $9/9$ & $18/18$ \\
Jacobian discrepancy plus reference allowance below trial bound & $9/9$ & $18/18$ \\
Position and orientation comparisons within their allowances & $9/9$ & $18/18$ \\
Available module-budget combinations & $9/27$ & $36/54$ \\
Available structured bounds below shared-input normwise bounds & $9/9$ & $36/36$ \\
Available bounds meeting illustrative $\delta_J=0.05$ & $9/9$ & $36/36$ \\
Stable numerical-reference repeats & $3/3$ & $6/6$ \\
Evaluable directional comparisons & $216/216$ & $431/432$ \\
\addlinespace[2pt]
Reconstruction median range (ms) & $1.029$--$1.444$ & $1.399$--$1.702$ \\
Bound-evaluation median range (ms) & $3.801$--$10.222$ & $3.965$--$6.344$ \\
\bottomrule
\end{tabular}
\endgroup
\end{table}

\subsection{Task-Jacobian results and module-budget screening}
\label{subsec:jacobian_results}
For the stored normalized Jacobian $\widehat{\widetilde{\boldsymbol J}}$ and numerical reference $\widetilde{\boldsymbol J}_{\mathit{ref}}$, the observed discrepancy is $d_{J,\mathrm{obs}}=\|\widehat{\widetilde{\boldsymbol J}}-\widetilde{\boldsymbol J}_{\mathit{ref}}\|_2$. The notation $\varepsilon_{J,\mathrm{prod}}$ denotes $\varepsilon_{J,\mathrm{tot}}$ evaluated at the trial reconstruction center, and $\varepsilon_{J,\mathit{ref}}$ denotes the corresponding reference-center allowance. Table~\ref{tab:global_trial_results} reports the trial comparisons. Every observed discrepancy lies below its allowance, with ratios ranging from approximately $1.58\times10^{-7}$ to $1.89\times10^{-5}$. All 27 comparisons also satisfy $d_{J,\mathrm{obs}}+\varepsilon_{J,\mathit{ref}}\leq\varepsilon_{J,\mathrm{prod}}$, and all nine reference repeats meet the stated agreement criteria. The separate computed-to-anchor term in~\eqref{eq:total_task_map_discrepancy} is zero under the nominal convention in these evaluations, so $\varepsilon_J=\varepsilon_{J,\mathrm{prod}}$ in the trial records. Table~\ref{tab:global_trial_results} also reports the anchor-based rank test~\eqref{eq:task_rank_margin} for the $6\times7$ normalized Jacobian. All 27 margins are positive, ranging from $0.220171$ to $0.248745$, with the smallest at C08 ($\sigma_{\min}=0.220184$, $\varepsilon_J\simeq1.316\times10^{-5}$). The 45 available hypothetical budget bounds likewise give positive rank margins, the smallest approximately $0.170399$. The other 36 budget cells lack the task allowance required for the test. These are nominal sufficient assessments of task rank, distinct from the local closure-rank gates.

\begin{table}[pos=!tb,abovecap=4pt,belowcap=1pt]
\caption{Normalized task-Jacobian discrepancies and conditional trial-center allowances for the nine configurations. The corrector tolerance is $f_g\,10^{-9}$, and all numerical columns are dimensionless. For each of C04--C09, the three tolerance settings give identical exported discrepancies and allowances, so one row represents all three trials. Ratios are calculated before rounding. The final two columns give the smallest singular value of the normalized anchor and its sufficient rank margin from the diagnostic replay using the matching structured allowance $\varepsilon_J$.}
\label{tab:configuration_coordinates}
\label{tab:global_trial_results}
\centering
\begingroup
\small
\setlength{\tabcolsep}{4.6pt}
\renewcommand{\arraystretch}{1.14}
\begin{tabular}{@{}ccrrrrr@{}}
\toprule
Case & $f_g$ & $d_{J,\mathrm{obs}}$ & $\varepsilon_{J,\mathrm{prod}}$ & $d_{J,\mathrm{obs}}/\varepsilon_{J,\mathrm{prod}}$ & $\sigma_{\min}(\widetilde{\boldsymbol J}_{\mathrm{anc}})$ & $\underline\sigma$ \\
\midrule
C01 & 0.1 & $5.656\times10^{-14}$ & $2.994\times10^{-9}$ & $1.89\times10^{-5}$ & $0.235023$ & $0.235023$ \\
C01 & 1 & $2.935\times10^{-9}$ & $7.055\times10^{-3}$ & $4.16\times10^{-7}$ & $0.235023$ & $0.227968$ \\
C01 & 10 & $2.935\times10^{-9}$ & $7.055\times10^{-3}$ & $4.16\times10^{-7}$ & $0.235023$ & $0.227968$ \\
\addlinespace[2pt]
C02 & 0.1 & $8.460\times10^{-15}$ & $2.570\times10^{-9}$ & $3.29\times10^{-6}$ & $0.234491$ & $0.234491$ \\
C02 & 1 & $2.935\times10^{-9}$ & $7.055\times10^{-3}$ & $4.16\times10^{-7}$ & $0.234491$ & $0.227435$ \\
C02 & 10 & $2.935\times10^{-9}$ & $7.055\times10^{-3}$ & $4.16\times10^{-7}$ & $0.234491$ & $0.227435$ \\
\addlinespace[2pt]
C03 & 0.1 & $6.195\times10^{-15}$ & $2.577\times10^{-9}$ & $2.40\times10^{-6}$ & $0.234619$ & $0.234619$ \\
C03 & 1 & $2.935\times10^{-9}$ & $7.055\times10^{-3}$ & $4.16\times10^{-7}$ & $0.234619$ & $0.227564$ \\
C03 & 10 & $6.780\times10^{-8}$ & $7.304\times10^{-3}$ & $9.28\times10^{-6}$ & $0.234619$ & $0.227315$ \\
\addlinespace[2pt]
C04 & $0.1,1,10$ & $1.010\times10^{-14}$ & $8.856\times10^{-9}$ & $1.14\times10^{-6}$ & $0.229241$ & $0.229241$ \\
C05 & $0.1,1,10$ & $4.138\times10^{-15}$ & $9.629\times10^{-9}$ & $4.30\times10^{-7}$ & $0.241695$ & $0.241695$ \\
C06 & $0.1,1,10$ & $5.385\times10^{-15}$ & $8.857\times10^{-9}$ & $6.08\times10^{-7}$ & $0.228104$ & $0.228104$ \\
C07 & $0.1,1,10$ & $1.519\times10^{-15}$ & $9.629\times10^{-9}$ & $1.58\times10^{-7}$ & $0.240628$ & $0.240628$ \\
C08 & $0.1,1,10$ & $5.573\times10^{-12}$ & $1.316\times10^{-5}$ & $4.23\times10^{-7}$ & $0.220184$ & $0.220171$ \\
C09 & $0.1,1,10$ & $3.429\times10^{-12}$ & $1.039\times10^{-5}$ & $3.30\times10^{-7}$ & $0.248755$ & $0.248745$ \\
\bottomrule
\end{tabular}
\endgroup
\end{table}

\begin{figure}[pos=!tb,abovecap=3pt,belowcap=1pt]
\centering
\includegraphics[width=0.72\columnwidth,trim={0cm 0cm 0cm 0cm},clip]{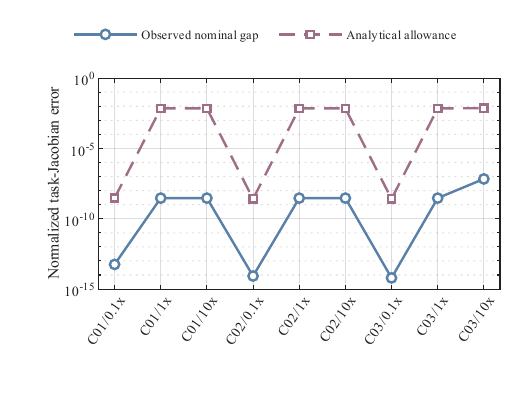}
\caption{Normalized task-Jacobian discrepancies and conditional allowances for C01--C03 at the three global corrector-tolerance multipliers. The repeated evaluation reproduces the plotted discrepancy and allowance values to the precision of the exported records. All nine observed discrepancies are below the corresponding nominally evaluated bounds.}
\label{fig:global_task_comparison}
\end{figure}

Figure~\ref{fig:global_task_comparison} shows the results for C01--C03. The identical discrepancies and allowances at $f_g=1$ and $10$ for C01 and C02 recur in this run. For each of C04--C09, all three tolerance settings give the same exported discrepancies and allowances; their rows are therefore combined in Table~\ref{tab:global_trial_results}. These settings do not produce three distinct observed accuracy levels at those centers. The large discrepancy-to-bound separation shows conservatism in the nominal Jacobian comparisons, particularly where the discrepancies approach numerical-reference resolution.

\begin{figure}[pos=!tb,abovecap=3pt,belowcap=1pt]
\centering
\includegraphics[width=0.95\textwidth,trim={0cm 1.0cm 0cm 0cm},clip]{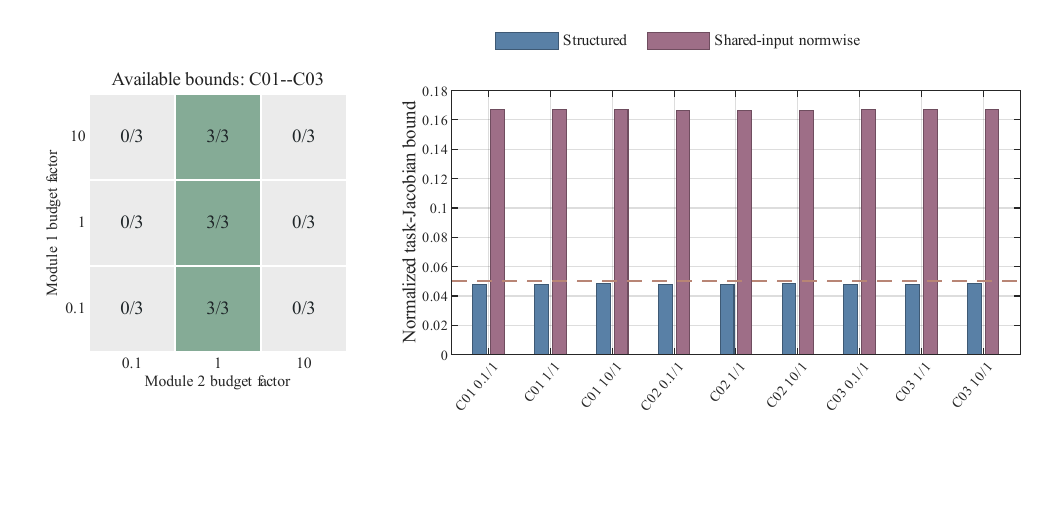}
\caption{Fixed-center module-budget assessment for C01--C03. \textit{Left}: each cell gives the number of available conditional bounds among the three configurations; zero counts retain the excluded combinations. \textit{Right}: structured and shared-input normwise allowances for the nine available combinations, labelled by configuration and $f_1/f_2$. The dashed line is the illustrative, post-hoc threshold $\delta_J=0.05$. Additional-case availability and exclusion causes are reported in the text.}
\label{fig:module_budget_availability}
\end{figure}

For the fixed-center budget assessment, 45 of 81 combinations have available bounds: nine of 27 for C01--C03 and 36 of 54 for C04--C09. Figure~\ref{fig:module_budget_availability} summarizes the original configurations with one shared availability map and compares their available structured and normwise bounds. For C01--C03, availability occurs only at $f_2=1$ for each Module 1 factor. For C04--C09, both $f_2=0.1$ and $1$ are available for every Module 1 factor. The 45 structured bounds range from approximately $0.004254$ to $0.049785$ and meet the illustrative, post-hoc threshold $\delta_J=0.05$. The normwise-to-structured ratios range from approximately $3.30$ to $5.24$; only 15 of the same 45 combinations satisfy the threshold with the shared-input normwise bound. Thus retaining componentwise structure enlarges the set accepted by this sufficient screen, using the same enclosure inputs.

The module diagnostics identify the two causes of unavailability. At the original centers, the Module 2 selected residual is approximately $2.612\times10^{-10}$, exceeding the hypothetical $10^{-10}$ budget at $f_2=0.1$; this excludes nine combinations. The additional centers have smaller Module 2 residuals and admit that budget. At $f_2=10$, the Module 2 radius search does not establish a self-mapping contraction for any of the nine centers, accounting for the other 27 unavailable combinations. Module 1 bounds are available throughout. Repeated evaluation of the diagnostics agrees with all 81 retained grid statuses and available bounds, and all 162 local statuses. This explains why a tighter budget can be incompatible with an unchanged center while a looser budget can exceed the range established by the sufficient-radius construction.

\begin{figure}[pos=!tb,abovecap=3pt,belowcap=1pt]
\centering
\includegraphics[width=0.95\textwidth,trim={0cm 0.4cm 0cm 0cm},clip]{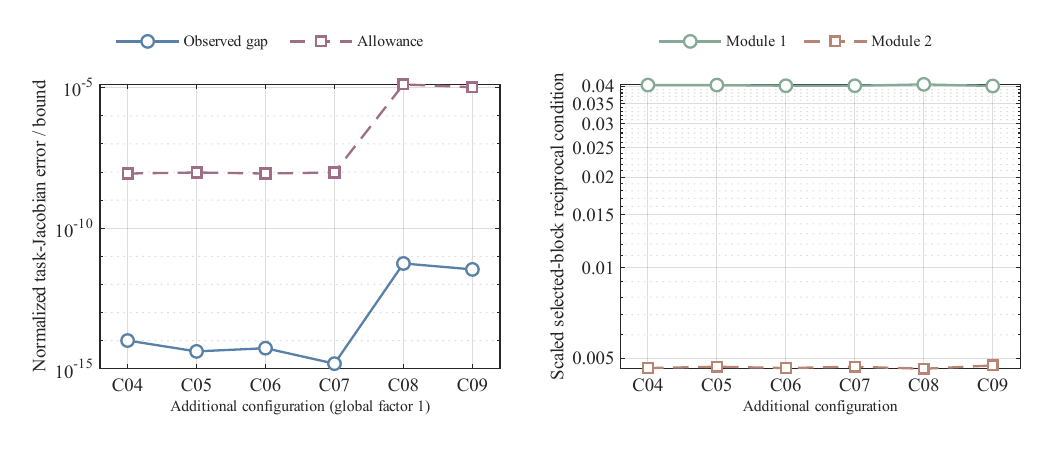}
\caption{Additional fixed configurations C04--C09. \textit{Left}: observed normalized task-Jacobian discrepancies and conditional allowances at $f_g=1$; each case has the same exported error quantities at the other two tolerance settings. \textit{Right}: reciprocal condition estimates of the scaled selected passive-Jacobian blocks at the baseline centers. Both modules change passive configuration, while their conditioning varies only modestly across the tested cases.}
\label{fig:additional_configuration_validation}
\end{figure}

Figure~\ref{fig:additional_configuration_validation} shows the additional-case comparisons and their module-wise conditioning. Across all nine baselines, the scaled selected passive-block reciprocal condition estimates range from approximately $0.040087$ to $0.040586$ for Module 1 and from $0.0046320$ to $0.0047499$ for Module 2. Their within-module maximum-to-minimum ratios are only $1.01245$ and $1.02546$, respectively. The extension therefore samples changes in both modules but remains a narrowly conditioned local assessment. In all 27 timed trials, bound evaluation exceeds reconstruction time, with paired ratios of approximately $2.72$--$9.75$. The timing ranges in Table~\ref{tab:host_three_case} characterize this cost. Further closure correction and a task-level error estimate provide complementary information: the former changes the reconstructed state, while the latter assesses its numerical accuracy for a specified task requirement.

\subsection{Contribution diagnostics and physical interpretation}
\label{subsec:contribution_diagnostics}
The contribution diagnostics from the replay distinguish the magnitude of each nonnegative term in~\eqref{eq:task_jacobian_allowance} from the corresponding signed changes observed against the numerical reference. In the common normalization, the separate contribution norms are
\[
 e_F=\|\SX\boldsymbol B_F|\Manc|\boldsymbol H\|_2,\quad
 e_M=\|\SX|\boldsymbol F_0|\boldsymbol B_M\boldsymbol H\|_2,\quad
 e_\times=\|\SX\boldsymbol B_F\boldsymbol B_M\boldsymbol H\|_2.
\]
The total $\varepsilon_J$ is the norm of the sum of the normalized nonnegative matrices, with $\max\{e_F,e_M,e_\times\}\leq\varepsilon_J\leq e_F+e_M+e_\times$. Across the 27 trial-center evaluations and the 45 available budget evaluations, $e_M/\varepsilon_J$ is between approximately $0.999992$ and $0.9999997$; $e_M$ exceeds $e_F$ by at least $5.4\times10^3$, and $e_\times/e_M$ is at most $1.15\times10^{-7}$. The module-specific norms $e_{M,m}$ are formed by restricting $\boldsymbol B_M$ to the passive rows of module $m$ in the expression for $e_M$. Module 2 gives the larger of these norms in all 27 trials and 39 of the 45 available budget cells. The six exceptions are C04--C09 at $(f_1,f_2)=(10,0.1)$, where the looser hypothetical Module 1 budget reverses the ordering.

For comparison with the nonnegative allowances, the signed factor differences are $\Delta\boldsymbol F_{\mathit{ref}}=\boldsymbol F(\boldsymbol q_{\mathit{ref}})-\boldsymbol F_0$ and $\Delta\boldsymbol M_{\mathit{ref}}=\boldsymbol M_{\mathit{ref}}-\Manc$, with the reference map assembled at the refined configuration. The norms of $\SX\Delta\boldsymbol F_{\mathit{ref}}\Manc\boldsymbol H$ and $\SX\boldsymbol F_0\Delta\boldsymbol M_{\mathit{ref}}\boldsymbol H$ measure the corresponding signed product changes. In the six trials for which the observed Jacobian discrepancy exceeds the nominal reference allowance, $e_M$ is approximately $(9.81\text{--}9.90)\times10^6$ times its signed reference-change norm, whereas $e_F$ is about $7.1$--$32.7$ times its corresponding norm. Thus the dominant bound term need not be the dominant observed change. The remaining trials are interpreted at the available reference resolution. These comparisons neither constitute additive error fractions nor determine the exact errors of the individual factors.

The local diagnostics further identify which ingredients of the computed-map bound in~\eqref{eq:direct_computed_map_allowance} govern its magnitude. For all 54 trial module evaluations, the preconditioned center-map correction contributes less than $3.81\times10^{-6}$ of the local map allowance, and the contraction factor $\theta$ is at most $0.01121$, giving an inverse factor below $1.012$. The preconditioned curvature $\Gamma_{pp}$ ranges from $2.71\times10^5$ to $3.10\times10^5$ in Module 1 and from $1.77\times10^7$ to $1.80\times10^7$ in Module 2. The computed-map sensitivity $\Gamma_{p\widehat D}$ is approximately 17--51 times $\Gamma_{pa}$. Table~\ref{tab:local_conservatism} presents the global contribution norms and local quantities for representative observed-residual trials at different discrepancy levels. Here $d_{p,m}$ and $d_{D,m}$ are the normalized infinity-norm discrepancies in the passive coordinates and differential map, respectively, relative to the refined numerical reference; $\eta_{D,\mathit{ref},m}$ is its local differential-map allowance.

\begin{table}[pos=!tb,abovecap=4pt,belowcap=1pt]
\caption{Global contribution norms (top) and local diagnostics (bottom) at representative trial centers using observed residuals. All quantities are dimensionless. A dagger identifies a numerical reference outside the nominal radius.}
\label{tab:local_conservatism}
\centering
\begingroup
\small
\setlength{\tabcolsep}{3.4pt}
\renewcommand{\arraystretch}{1.12}
\begin{tabular}{@{}crrrrrr@{}}
\toprule
Case ($f_g$) & $e_F$ & $e_M$ & $e_\times$ & $\varepsilon_J$ & $e_{M,1}$ & $e_{M,2}$ \\
\midrule
C03 (10) & $4.81\times10^{-7}$ & $7.30\times10^{-3}$ & $1.70\times10^{-10}$ & $7.30\times10^{-3}$ & $1.97\times10^{-3}$ & $7.06\times10^{-3}$ \\
C04 (1) & $1.42\times10^{-13}$ & $8.86\times10^{-9}$ & $4.37\times10^{-23}$ & $8.86\times10^{-9}$ & $8.16\times10^{-11}$ & $8.86\times10^{-9}$ \\
C08 (1) & $2.07\times10^{-10}$ & $1.32\times10^{-5}$ & $8.97\times10^{-17}$ & $1.32\times10^{-5}$ & $6.46\times10^{-8}$ & $1.32\times10^{-5}$ \\
\bottomrule
\end{tabular}
\par\vspace{6pt}
\begin{tabular}{@{}ccrrrrrr@{}}
\toprule
Case ($f_g$) & $m$ & $\varsigma_m$ & $d_{p,m}/\varsigma_m$ & $\theta$ & $\varepsilon_{D,\mathrm{anc},m}$ & $d_{D,m}$ & $\eta_{D,\mathit{ref},m}$ \\
\midrule
C03 (10) & 1 & $5.89\times10^{-9}$ & $0.999165$ & $1.67\times10^{-3}$ & $1.00\times10^{-3}$ & $1.05\times10^{-10}$ & $4.80\times10^{-12}$ \\
C03 (10) & 2 & $6.27\times10^{-10}$ & $0.994392$ & $1.12\times10^{-2}$ & $1.53\times10^{-3}$ & $3.92\times10^{-10}$ & $4.38\times10^{-11}$ \\
C04 (1) & 1 & $2.29\times10^{-16}$ & $0.740240$ & $6.82\times10^{-11}$ & $4.13\times10^{-11}$ & $2.22\times10^{-16}$ & $1.53\times10^{-16}$ \\
C04 (1) & 2 & $7.90\times10^{-16}$ & $1.028985^{\dagger}$ & $1.42\times10^{-8}$ & $1.92\times10^{-9}$ & $1.33\times10^{-15}$ & $1.81\times10^{-10}$ \\
C08 (1) & 1 & $1.71\times10^{-13}$ & $1.000006^{\dagger}$ & $5.30\times10^{-8}$ & $3.25\times10^{-8}$ & $1.36\times10^{-12}$ & $2.69\times10^{-12}$ \\
C08 (1) & 2 & $1.16\times10^{-12}$ & $0.996986$ & $2.09\times10^{-5}$ & $2.85\times10^{-6}$ & $1.30\times10^{-12}$ & $3.55\times10^{-10}$ \\
\bottomrule
\end{tabular}
\endgroup
\end{table}

For C03 at $f_g=10$, both numerical-reference passive discrepancies are within 1\% of their respective radii and both references lie inside the nominal boxes. Nevertheless, $\varsigma_m\Gamma_{pa}$ exceeds the sampled preconditioned active-block change by factors of approximately $2.13\times10^4$ and $1.03\times10^6$ for Modules 1 and 2, respectively. For the corresponding preconditioned passive-block change acting on $\Dhat$, $\varsigma_m\Gamma_{p\widehat D}$ exceeds the sampled change by approximately $3.91\times10^5$ and $3.51\times10^6$. The derivative comparisons use the same center preconditioner and residual scaling at both points. These results point to the uniform preconditioned derivative majorants as the principal source of the observed local overestimation in this case, rather than amplification by the inverse factor or a large center-map correction. The diagnostics do not isolate the individual path or logarithm inequalities responsible. Since each local map has a single column, its induced infinity norm is its largest absolute entry, so no multi-column row sum is involved in the conversion~\eqref{eq:physical_differential_allowance}; assigning the common scalar allowance to every passive row can still introduce conservatism.

All 54 nominal trial module map comparisons satisfy $d_{D,m}+\eta_{D,\mathit{ref},m}\leq\varepsilon_{D,\mathrm{anc},m}$. In contrast, 18 numerical-reference passive discrepancies exceed their nominal radii by at most $4.89\times10^{-16}$ in normalized coordinates, and 29 passive comparisons have negative margins after adding the reference-center radius (maximum deficit $5.03\times10^{-16}$). The small absolute differences highlight the reference-resolution and finite-arithmetic limitations of these local comparisons. They are not tests against a known exact root. In the fixed-center budget study, the derivative and coordinate differences for a given configuration are unchanged as the hypothetical budgets vary, whereas the associated radii and majorants depend on the budget. The corresponding gaps characterize the hypothetical allowances at fixed centers.

As a physical interpretation of the trial bounds, consider an illustrative input limit $\alpha=0.1\ \mathrm{s}^{-1}$ with $\|\boldsymbol H^{-1}\boldsymbol u\|_2\leq\alpha$, corresponding to one tenth of the declared active-coordinate spans per second in the scaled Euclidean norm. The largest trial-center total allowance is bounded by the rounded-up value $\bar\varepsilon=7.305\times10^{-3}$. Corollary~\ref{cor:task_rate} then gives the following common same-input rate-discrepancy allowances for the sampled centers, in physical units:
\[
 e_{v,\mathrm{mm/s}}=0.7305\frac{\chi_r}{1\ \mathrm{m}}\ \mathrm{mm/s},\qquad
 e_{\omega,\mathrm{rad/s}}=7.305\times10^{-4}\frac{\chi_R}{1\ \mathrm{rad}}\ \mathrm{rad/s}.
\]
For illustrative rate-error limits of $1$ mm/s and $0.001$ rad/s under that input limit, the sufficient Jacobian budget is $\delta_J=0.01/\max\{\chi_r/(1\ \mathrm{m}),\chi_R/(1\ \mathrm{rad})\}$. The calculated trial allowances would satisfy both limits if $\bar\varepsilon\leq\delta_J$, equivalently if the larger scale ratio is at most $0.01/\bar\varepsilon\simeq1.369$. This scale-explicit example retains the actual normalization; the withheld scale magnitudes preclude reporting an evaluated physical pass/fail result. The original $0.05$ threshold is a separate illustrative normalized operator budget, not a percentage of the task velocity. The interpretation concerns the same supplied input at each sampled configuration; a desired-rate comparison additionally includes the computed target-rate residual in~\eqref{eq:task_budget_test}.

\subsection{Supporting end-point pose results}
\label{subsec:pose_results}
The position discrepancy from the numerical reference is $d_{r,\mathrm{obs}}=10^3\|\widehat{\boldsymbol r}-\boldsymbol r_{\mathit{ref}}\|_2$ in mm, and the orientation discrepancy is $d_{R,\mathrm{obs}}=d_{\SO(3)}(\widehat{\rot},\rot_{\mathit{ref}})$ in rad. The latter uses the shortest relative rotation angle of Corollary~\ref{cor:static_pose}, obtained through the Frobenius-to-angle relation. Table~\ref{tab:endpoint_pose_results} reports per-configuration maxima of these discrepancies and the associated nominal pose allowances over the three global trials.

\begin{table}[pos=!tb,abovecap=4pt,belowcap=1pt]
\caption{Supporting end-point pose comparisons in physical units. Each entry is the separate maximum over $f_g\in\{0.1,1,10\}$ at the stated configuration. Position discrepancies and allowances are in mm; orientation discrepancies and allowances are shortest-angle quantities in rad. Each of the 27 underlying trials was checked against its own allowance, rather than by comparing the maxima alone.}
\label{tab:endpoint_pose_results}
\centering
\begingroup
\small
\setlength{\tabcolsep}{8pt}
\renewcommand{\arraystretch}{1.14}
\begin{tabular}{@{}crrrr@{}}
\toprule
& \multicolumn{2}{c}{Position (mm)} & \multicolumn{2}{c}{Orientation (rad)} \\
\cmidrule(lr){2-3}\cmidrule(l){4-5}
Case & $\max d_{r,\mathrm{obs}}$ & $\max 10^3\varepsilon_r$ & $\max d_{R,\mathrm{obs}}$ & $\max\varepsilon_R$ \\
\midrule
C01 & $1.436\times10^{-6}$ & $1.767\times10^{-5}$ & $5.208\times10^{-10}$ & $8.250\times10^{-9}$ \\
C02 & $1.436\times10^{-6}$ & $1.751\times10^{-5}$ & $5.208\times10^{-10}$ & $8.177\times10^{-9}$ \\
C03 & $4.780\times10^{-5}$ & $1.045\times10^{-4}$ & $1.203\times10^{-8}$ & $3.706\times10^{-8}$ \\
\addlinespace[2pt]
C04 & $4.311\times10^{-12}$ & $2.614\times10^{-11}$ & $1.801\times10^{-15}$ & $1.174\times10^{-14}$ \\
C05 & $1.790\times10^{-12}$ & $2.787\times10^{-11}$ & $7.742\times10^{-16}$ & $1.255\times10^{-14}$ \\
C06 & $1.665\times10^{-12}$ & $2.918\times10^{-11}$ & $9.380\times10^{-16}$ & $1.271\times10^{-14}$ \\
C07 & $1.776\times10^{-12}$ & $3.083\times10^{-11}$ & $2.355\times10^{-16}$ & $1.349\times10^{-14}$ \\
C08 & $2.254\times10^{-9}$ & $3.676\times10^{-8}$ & $9.324\times10^{-13}$ & $1.680\times10^{-11}$ \\
C09 & $1.415\times10^{-9}$ & $2.693\times10^{-8}$ & $5.641\times10^{-13}$ & $1.237\times10^{-11}$ \\
\bottomrule
\end{tabular}
\endgroup
\end{table}

Each of the 27 position comparisons and 27 orientation comparisons lies below its own allowance. The largest paired discrepancy-to-allowance ratios are $0.538$ for position and $0.386$ for orientation, much larger than the corresponding Jacobian ratios. The position and orientation comparisons also retain positive nominal margins after adding the respective reference-center allowances. The largest trial allowances are $1.045\times10^{-4}$ mm and $3.706\times10^{-8}$ rad. These results give physical-unit measures of reconstruction error in the nominal model; they complement the primary Jacobian criterion without representing measured total robot positioning accuracy.

\subsection{Directional-Jacobian check}
\label{subsec:directional_validation}
A separate numerical-differentiation check compares the analytical reduced Jacobian with changes in reconstructed end-point poses. At each refined reference configuration, the independent coordinates are perturbed as $\boldsymbol a_\pm=\boldsymbol a\pm h\boldsymbol H\boldsymbol d$, where $\|\boldsymbol d\|_2=1$ and $h$ is dimensionless. The eight directions comprise the seven normalized coordinate axes and the mixed direction $d_j=(-1)^{j-1}/\sqrt7$. Nine step sizes are used: $10^{-2}$, $3\times10^{-3}$, $10^{-3}$, $3\times10^{-4}$, $10^{-4}$, $3\times10^{-5}$, $10^{-5}$, $3\times10^{-6}$, and $10^{-6}$. Both perturbed reconstructions start from the same refined center, with homotopy fallback disabled and the center's selected physical rows retained for comparison.

\begin{figure}[pos=!tb,abovecap=3pt,belowcap=1pt]
\centering
\includegraphics[width=0.95\textwidth,trim={0cm 0.6cm 0cm 0cm},clip]{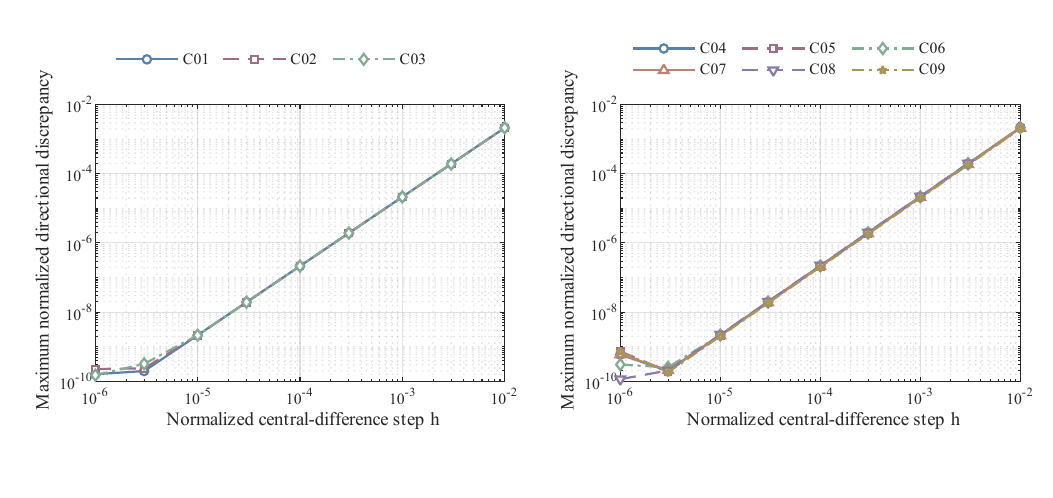}
\caption{Maximum normalized directional discrepancy $\max_{\boldsymbol d}e_h(\boldsymbol d)$ between the analytical Jacobian action and central differences of refined end-point poses. \textit{Left}: C01--C03. \textit{Right}: C04--C09. Each plotted value requires all eight directions at that step size. The C08 value at $h=10^{-2}$ is omitted under the ordered-selector comparison criterion because the negative $q_7$ perturbation permutes the selected closure rows. The decrease at intermediate steps and the smallest-step resolution effects are interpreted as nominal numerical comparisons.}
\label{fig:directional_jacobian_check}
\end{figure}

For the refined side poses $(\boldsymbol r_\pm,\rot_\pm)$, the translation-first normalized difference and its discrepancy are
\[
 \boldsymbol g_h(\boldsymbol d)=\frac{1}{2h}\SX\col\!\left(\boldsymbol r_+-\boldsymbol r_-,\Log(\rot_+\rot_-^{\mathsf T})^\vee\right),
 \qquad e_h(\boldsymbol d)=\|\boldsymbol g_h(\boldsymbol d)-\widetilde{\boldsymbol J}_{\mathit{ref}}\boldsymbol d\|_2.
\]
The relative rotation in this expression gives world-frame angular components, consistent with the analytical map. The pose products are evaluated separately without using the analytical task-derivative routine, while the nominal model and closure evaluator are shared.

Of the 648 planned direction--step comparisons, 647 are evaluable. The remaining comparison, the $q_7$ direction at C08 with $h=10^{-2}$, is excluded under the original fixed-order protocol. A targeted replay of the two perturbations shows that both direct reconstructions are accepted. The negative perturbation changes the pivoted-QR ordering of the Module 2 selector from $(9,3,11,5,4,10)$ to $(9,3,10,5,11,4)$, preserving the row set; the positive perturbation and Module 1 selectors retain their original order. With the center residual scaling, the reciprocal condition estimate of the original Module 2 block is approximately $0.0046320$ at the center, $0.0046257$ at the positive side, and $0.0046394$ at the negative side. The passive blocks retain their full column ranks and satisfy the recorded regularity gates on both sides. The exclusion thus reflects a row permutation, with no substantial conditioning deterioration observed; the original 647/648 comparison count is retained. Figure~\ref{fig:directional_jacobian_check} reports each case's maximum over all eight directions only when the full set is available. Across all cases and directions, the maximum discrepancy decreases from $2.23\times10^{-5}$ at $h=10^{-3}$ to $2.23\times10^{-9}$ at $h=10^{-5}$, consistent with second-order step-size behavior. It reaches $3.19\times10^{-10}$ at $h=3\times10^{-6}$ and increases to $7.28\times10^{-10}$ at $h=10^{-6}$, indicating sensitivity to numerical resolution at the smallest steps. For the evaluable comparisons, the side-reference raw full-closure residuals are at most $1.15\times10^{-15}$ and the remaining scaled Newton-correction diagnostics at most $4.20\times10^{-16}$. Together, the step-size trend and recorded closure diagnostics support the analytical differential calculation through a separate numerical differentiation route.

\FloatBarrier

\section{Conclusion}
\label{sec:conclusion}
This work establishes a conditional connection between module-local closure reconstruction and the normalized reduced task Jacobian within the path assembly framework. The local analysis combines residual allowances, inverse bounds, and derivative variation to bound passive-coordinate and differential-map errors, including computed-center corrections. Propagation through the ordered paths preserves module support, exact independent-coordinate rows, and nonlinear cross terms. The resulting dimensionless allowance gives sufficient module-specific accuracy requirements for a prescribed task-Jacobian budget, while static end-point pose and same-input instantaneous task-rate bounds provide complementary physical interpretations.

The nominal-double assessment at nine fixed configurations supports this connection through reconstruction trials, refined numerical references, and a separate directional-differentiation check. All observed Jacobian and supporting pose discrepancies remain within their corresponding allowances, and the reference-centered margins for these task-level comparisons are positive. Of the 81 hypothetical module-budget combinations, 45 yield available bounds; all 45 structured bounds meet the illustrative, post-hoc threshold of $0.05$, compared with 15 of the same combinations under the shared-input normwise estimate. This comparison shows how retaining componentwise structure can enlarge the set accepted by a sufficient accuracy screen. The sufficient anchor-based task-rank margins are positive for the evaluated trial and available budget bounds. The directional check exhibits the expected central-difference step-size trend over an intermediate range, with one comparison excluded under the ordered-selector criterion after a row permutation.

The contribution diagnostics identify differential-map variation as the dominant term of the conservative Jacobian allowances, with representative local comparisons pointing to the uniform derivative majorants. Passive-radius comparisons at numerical resolution illustrate the limitations of nominal evaluations. The conditioning range remains narrow, and bound evaluation takes longer than reconstruction in every measured trial. The scale-dependent rate interpretation connects the normalized allowance to physical input and task-error specifications. Its use to guide solver stopping requires a justified residual budget and reevaluation at each reconstruction center. Within the stated branch and domain conditions, the framework provides a systematic basis for relating modular reconstruction effort to task-kinematic accuracy.

\appendix
\makeatletter
\renewcommand{\@seccntformat}[1]{%
  \ifstrequal{#1}{section}{Appendix~}{}%
  \csname the#1\endcsname\@seccntDot\hskip 0.5em}
\makeatother
\section{Computed Quantities and Numerical Validity Conditions}
\label{app:numerical}
The computed-center bounds distinguish errors in the evaluated closure Jacobians from arithmetic errors in the map residual. For $b\in\{a,p\}$, $\widehat{\widetilde{\boldsymbol R}}_{b,0}$ approximates the exact normalized center block $\widetilde{\boldsymbol R}_{b,0}$, with nonnegative componentwise allowance $\boldsymbol E_b$ satisfying $|\widetilde{\boldsymbol R}_{b,0}-\widehat{\widetilde{\boldsymbol R}}_{b,0}|\leq\boldsymbol E_b$. The matrix $\widehat{\boldsymbol r}_D$ is the computed residual formed from these approximate blocks and the stored map $\Dhat$, while $\boldsymbol E_r$ bounds the componentwise arithmetic error of that evaluation. The exact-block residual $\boldsymbol r_D$ from~\eqref{eq:computed_differential_map_bound} therefore satisfies
\begin{equation}
 |\boldsymbol r_D|\leq|\widehat{\boldsymbol r}_D|+\boldsymbol E_r+
 \boldsymbol E_p|\Dhat|+\boldsymbol E_a.
 \label{eq:computed_block_correction}
\end{equation}
This supplies the center correction in~\eqref{eq:computed_differential_map_bound}, accounting for logarithm and derivative evaluation errors as well as the linear solve. For the approximate-inverse construction, the notation $\widehat{\boldsymbol A}_0=\widehat{\widetilde{\boldsymbol R}}_{p,0}$ and $\widehat{\boldsymbol C}_0=\widehat{\widetilde{\boldsymbol R}}_{a,0}$ denotes the computed passive and active blocks. The computed residual $\widehat{\widetilde{\boldsymbol c}}_0$ has a nonnegative allowance $\boldsymbol E_c$ satisfying $|\widetilde{\boldsymbol c}_0-\widehat{\widetilde{\boldsymbol c}}_0|\leq\boldsymbol E_c$, and $\widehat{\K}$ is a declared nonsingular approximation to the inverse passive block. The following quantities bound the preconditioned exact-center residual and the inverse defect, respectively:
\begin{equation}
 \eta_{\mathrm{obs}}=\|\widehat{\K}\widehat{\widetilde{\boldsymbol c}}_0\|_\infty+
 \||\widehat{\K}|\boldsymbol E_c\|_\infty,\qquad \delta_K=\|\boldsymbol I-\widehat{\K}\widehat{\boldsymbol A}_0\|_\infty+
 \||\widehat{\K}|\boldsymbol E_p\|_\infty.
 \label{eq:computed_center_preconditioner_gates}
\end{equation}
The residual bound $\eta$ may be taken as $\eta_{\mathrm{obs}}$ or obtained from a separately justified exact-residual budget, with evaluation uncertainty included once. The computed map residual evaluates $\widehat{\boldsymbol A}_0\Dhat+\widehat{\boldsymbol C}_0$, with its arithmetic uncertainty retained through $\boldsymbol E_r$. The allowances $\Gamma_{pa}$ and $\Gamma_{p\widehat D}$ follow~\eqref{eq:differential_curvature_allowances} using $\widehat{\K}$ and $\Dhat$ in place of $\K$ and $\Dzero$; here $\Gamma_{pp}$ likewise bounds the passive-Jacobian variation preconditioned by $\widehat{\K}$ on the same box. The resulting direct computed-map bound is
\begin{samepage}
\begin{subequations}
\label{eq:direct_computed_map_allowance}
\begin{equation}
 T_c:=\|\widehat{\K}\widehat{\boldsymbol r}_D\|_\infty
 +\||\widehat{\K}|(\boldsymbol E_r+\boldsymbol E_p|\Dhat|+\boldsymbol E_a)\|_\infty,
\end{equation}
\begin{equation}
 \|\Dstar-\Dhat\|_\infty
 \leq\frac{T_c+\varsigma(\Gamma_{pa}+\Gamma_{p\widehat D})}{1-\theta},
 \qquad \theta=\delta_K+\Gamma_{pp}\varsigma<1.
\end{equation}
\end{subequations}
\end{samepage}
The quantity $T_c$ includes the preconditioned center-map discrepancy and its evaluation allowances, while $\theta$ bounds the deviation of the preconditioned passive Jacobian from the identity throughout the box. Thus the center-map correction is already included in the total bound. Taking the separate arithmetic allowance $\boldsymbol E_r$ as zero without an enclosure uses the nominal exact-arithmetic convention of Subsection~\ref{subsec:numerical_conventions}.

For the corresponding radius test, $\Gamma_K$ denotes a uniform curvature allowance satisfying $\|\widehat{\K}D\Rp(\widetilde{\boldsymbol p})[\boldsymbol h]\|_\infty\leq\Gamma_K\|\boldsymbol h\|_\infty$ on $\mathcal B(\varsigma)$. It has the same role as $\Gamma_{pp}$ above, with the approximate preconditioner indicated explicitly. Retaining the center defect $\delta_K$ gives the sufficient conditions
\begin{equation}
 \eta+\delta_K\varsigma+\tfrac12\Gamma_K\varsigma^2\leq \varsigma,
 \qquad\theta=\delta_K+\Gamma_K\varsigma<1,
 \label{eq:approx_inverse_gate}
\end{equation}
where $\eta$ bounds the preconditioned exact-center residual, either by the observed plus uncertainty route or by an exact-residual budget. The proof of Lemma~\ref{lem:local_reconstruction} applies with this linear term and the stated input-error enclosures.

The radius test in Lemma~\ref{lem:local_reconstruction} requires a nonpositive self-map residual $Y+\Gamma_{pp}\varsigma^2/2-\varsigma$, together with the contraction and domain conditions. A search that reaches its computational work limit before establishing these inequalities leaves the sufficient bound unavailable and the existence of a root unresolved by this test. In the numerical assessment, the approximate-inverse radius iteration starts at $\eta/(1-\delta_K)$, tests at most 80 candidates, and requires a strictly interior passive box. If the next candidate does not increase the radius, the increment is the larger of $16\,\mathrm{eps}(\max(1,\varsigma))$ and $10^{-9}\max(\varsigma,\eta)$, where $\mathrm{eps}$ is the floating-point spacing in MATLAB. The implemented tests retain the strict contraction condition and nonpositive self-map residual.

The scalar majorant $e_J(s)$ in~\eqref{eq:closure_log_majorants} and its derivative $e'_J(s)$ are evaluated from positive series for the nonnegative normalized argument $s$. Truncation of $e_J$ after order $n\geq1$ leaves the real-arithmetic tail
\begin{equation}
 0\leq e_J(s)-\sum_{k=1}^n\frac{s^k}{(k+1)!}
 \leq\frac{s^{n+1}}{(n+2)!}\frac{1}{1-s/(n+3)},\qquad 0\leq s<n+3.
 \label{eq:series_tail}
\end{equation}
Successive positive terms have ratio at most $s/(n+3)$. For the derivative series the first omitted term is $(n+1)s^n/(n+2)!$ and a valid ratio bound is $(n+2)s/((n+1)(n+3))$, required to be below one. These bounds enclose analytical-series truncation. The numerical assessment uses 16 series terms with the positive geometric tails. The logarithm-domain supersolution iteration of Appendix~\ref{app:path_majorants} allows at most 200 iterations, with a stagnation threshold $10^{-13}\max(1,s_{\mathrm{next}})$ for the next scalar iterate $s_{\mathrm{next}}$. The subsequent enlargement starts with $\max(10^{-14},10^{-10}\max(1,s))$ and allows at most 48 doubling attempts, retaining self-consistency and the strict domain $e_J(s)<1$ below the positive root of $\exp(s)=1+2s$. These work limits govern the numerical search. Rounding effects and center discrepancies arising from different left-Jacobian implementations enter through the corresponding evaluation-error allowances.

\section{Path and Logarithm Majorants}
\label{app:path_majorants}
Starting from the path-derivative construction in~\cite{dastranj2026}, we derive explicit neighborhood bounds for fixed transforms and constant-twist motion factors. The construction below supplies the majorants used in Section~\ref{sec:local}; other transformation models require corresponding derivative bounds valid over the stated neighborhood.

\subsection{Prefix derivatives, signed scales, and repeated coordinates}
In the ordered path $\boldsymbol T=\prod_i\boldsymbol U_i$, the index $i$ identifies a transformation factor. A motion factor has the form $\boldsymbol U_i=\exp(\boldsymbol\xi_i^\wedge(q_{j(i)}-o_i))$, where $j(i)$ selects its global coordinate, $o_i$ is its fixed offset, and $\boldsymbol\xi_i$ is its constant angular-first twist. The wedge operator maps a twist to its Lie-algebra matrix, and $\vee$ is the inverse operation. The signed factor scale is included in $\boldsymbol\xi_i$, whereas differentiation with respect to a normalized coordinate introduces the positive coordinate scale $d_j$ from Subsection~\ref{subsec:numerical_conventions}. Fixed factors have zero derivative but affect subsequent path prefixes.

The auxiliary normalization $\boldsymbol N_\rho=\diag(\boldsymbol I_3,L^{-1}\boldsymbol I_3)$ uses a declared positive length $L$, with twist norm $\|\boldsymbol x\|_\rho=\|\boldsymbol N_\rho\boldsymbol x\|_\infty$ and its induced matrix norm. Here $\Ad_{\boldsymbol T}$ denotes the rigid-transform adjoint on twists, while $\ad_{\boldsymbol x}$ maps a twist to its Lie bracket with $\boldsymbol x$. Consistent translation scaling preserves the bracket relation, giving $\|\ad_{\boldsymbol x}\|_\infty\leq4\|\boldsymbol x\|_\infty$ when $\boldsymbol x$ is expressed in the normalized coordinates. A prefix whose translation norm is bounded by the scalar $P$ has normalized adjoint norm at most $\sqrt3(1+2P/L)$: the rotation block has infinity norm at most $\sqrt3$, and the normalized skew-translation factor has norm at most $2P/L$.

A fixed factor contributes its translation norm to a path translation bound. A motion factor contributes at most $\max|q_{j(i)}-o_i|\|\boldsymbol\xi_{i,v}\|_2$ on the box, where $\boldsymbol\xi_{i,v}$ is the linear block of the twist. Rotation orthogonality permits summing these contributions. One admissible choice of $L$ is the larger translation bound of the two paths defining a closure element, provided this length is positive; a zero value in a pure-rotation case leaves this particular choice unavailable.

The prefix $\boldsymbol A_i=\prod_{h<i}\boldsymbol U_h$ contains the factors preceding $i$, with $\overline A_i$ bounding its normalized adjoint norm uniformly on the box. In this appendix, $\partial_j$ denotes differentiation with respect to the normalized coordinate $\widetilde q_j$. The corresponding spatial column $\boldsymbol\eta_j$, prefix allowance $E_{<i,k}$, column allowance $E_j$, and derivative allowance $H_{jk}$ are
\begin{subequations}
\label{eq:prefix_majorants}
\begin{equation}
 \boldsymbol\eta_j=\sum_{i:j(i)=j}\Ad_{\boldsymbol A_i}\boldsymbol\xi_i d_j,
 \qquad E_{<i,k}=\sum_{h<i:j(h)=k}\overline A_h\|\boldsymbol\xi_h d_k\|_\rho,
\end{equation}
\begin{equation}
 E_j=\sum_{i:j(i)=j}\overline A_i\|\boldsymbol\xi_i d_j\|_\rho,
 \qquad H_{jk}=4\sum_{i:j(i)=j}E_{<i,k}\overline A_i\|\boldsymbol\xi_i d_j\|_\rho.
\end{equation}
\end{subequations}
The quantity $E_{<i,k}$ bounds the spatial variation of prefix $\boldsymbol A_i$ with respect to $\widetilde q_k$. The resulting bounds are $\|\boldsymbol\eta_j\|_\rho\leq E_j$ and $\|\partial_k\boldsymbol\eta_j\|_\rho\leq H_{jk}$; the latter follows by differentiating the adjoint and bounding the resulting Lie bracket. The sums retain every occurrence of a coordinate along the path.

\subsection{Closure transport and logarithm variation}
For one closure element, its index is suppressed and superscripts $+$ and $-$ abbreviate the two path labels $\langle+\rangle$ and $\langle-\rangle$. The matrix $\boldsymbol Q_c=\Ad_{(\boldsymbol T^-)^{-1}}$ transports spatial columns into the comparison frame. Thus $\boldsymbol\zeta_j=\boldsymbol Q_c(\boldsymbol\eta_j^+-\boldsymbol\eta_j^-)$ is the transported path difference, and $\boldsymbol w_k=\boldsymbol Q_c\boldsymbol\eta_k^-$ describes variation of the transporting path. With $\boldsymbol J_l(\boldsymbol\rho)$ denoting the left Jacobian of the exponential map on $\SE(3)$, the closure and transport derivatives satisfy~\cite{dastranj2026,sola2021micro}
\begin{equation}
 \partial_j\boldsymbol\rho=\boldsymbol J_l(\boldsymbol\rho)^{-1}\boldsymbol\zeta_j,\qquad \partial_k\boldsymbol\zeta_j=-\ad_{\boldsymbol w_k}\boldsymbol\zeta_j+
 \boldsymbol Q_c(\partial_k\boldsymbol\eta_j^+-\partial_k\boldsymbol\eta_j^-).
 \label{eq:closure_transport_derivative}
\end{equation}
These relations may equivalently be expressed in the auxiliary normalized coordinates. The inverse path has the same translation norm as the original path, so the preceding adjoint estimate also supplies a uniform bound $\overline Q_c$ for $\boldsymbol Q_c$. With $E_j^\pm$ and $H_{jk}^\pm$ obtained for the respective paths, the quantities $Z_j=\overline Q_c(E_j^++E_j^-)$, $W_k=\overline Q_cE_k^-$, and $Z_{jk}=4W_kZ_j+\overline Q_c(H_{jk}^++H_{jk}^-)$ bound $\|\boldsymbol\zeta_j\|_\rho$, $\|\boldsymbol w_k\|_\rho$, and $\|\partial_k\boldsymbol\zeta_j\|_\rho$, respectively.

For a nonnegative bound $s$ on the normalized logarithmic adjoint norm, the left-Jacobian series~\cite{barfoot2017state} and its derivative give the scalar majorants
\begin{equation}
 e_J(s)=\sum_{k\geq1}\frac{s^k}{(k+1)!},\qquad M_J(s)=(1-e_J(s))^{-1},\qquad K_J(s)=4M_J(s)^2e'_J(s),\qquad e_J(s)<1.
 \label{eq:closure_log_majorants}
\end{equation}
Here $e_J$ bounds the deviation of the left Jacobian from the identity, $M_J$ bounds its inverse, and $K_J$ bounds variation of the inverse with respect to a normalized logarithmic direction; $e'_J$ denotes differentiation with respect to $s$. Their continuous values at zero are $e_J(0)=0$, $M_J(0)=1$, and $K_J(0)=2$. The estimates follow from $\boldsymbol J_l=\sum_{k\geq0}\ad_{\boldsymbol\rho}^k/(k+1)!$, a Neumann inverse, and differentiation of that inverse. The factor $4$ converts a direction norm into an adjoint norm.

On the passive box $\mathcal B(\varsigma)$, $\boldsymbol\rho_0$ is the center logarithm and $I_p$ indexes the passive coordinates that vary. A bound $s_\rho$ on $4\|\boldsymbol N_\rho\boldsymbol\rho\|_\infty$, and hence on the normalized logarithmic adjoint norm, is established by
\begin{equation}
 4\left(\|\boldsymbol N_\rho\boldsymbol\rho_0\|_\infty+
 M_J(s_\rho)\varsigma\sum_{k\in I_p}Z_k\right)\leq s_\rho,\qquad e_J(s_\rho)<1.
 \label{eq:closure_log_domain_test}
\end{equation}
Integration of the logarithmic differential along a segment from the center bounds its variation by the radius-dependent term in~\eqref{eq:closure_log_domain_test}. A first crossing of $s_\rho$ strictly before the segment endpoint would have a smaller positive accumulated variation than the full-radius bound, contradicting that inequality. If the variation bound vanishes, the logarithm is constant. The argument therefore bounds the normalized logarithm throughout the box under the independently stated smooth-log-domain assumption.

The inverse left-Jacobian function $\mathcal L(\boldsymbol\rho)=\boldsymbol J_l(\boldsymbol\rho)^{-1}$ has directional derivative $D\mathcal L[\cdot]$. With $M_J$ and $K_J$ evaluated at $s_\rho$, differentiation gives the scalar second-derivative allowance $h_{jk}$ and its componentwise counterpart:
\begin{equation}
 \partial_k\partial_j\boldsymbol\rho
 =D\mathcal L[\mathcal L\boldsymbol\zeta_k]\boldsymbol\zeta_j+
 \mathcal L\partial_k\boldsymbol\zeta_j,\qquad h_{jk}=K_JM_JZ_kZ_j+M_JZ_{jk},\qquad |\partial_k\partial_j\boldsymbol\rho|\leq\boldsymbol N_\rho^{-1}\boldsymbol1_6h_{jk}.
 \label{eq:closure_hessian_columns}
\end{equation}
Here $\boldsymbol1_6$ is the six-entry all-ones vector. Stacking the column allowances over the closure elements and applying $\boldsymbol S_m$ and $\boldsymbol\Lambda_{m,c}^{-1}$ gives the residual-Hessian bounds $\boldsymbol{\mathcal H}^{pb}_{m,k}$ introduced in Subsection~\ref{subsec:differential_map_sensitivity}, with the coordinate scales already included. Consequently, $\Gamma_{pp}=\sum_k\||\K|\boldsymbol{\mathcal H}^{pp}_{k}\|_\infty$ satisfies~\eqref{eq:preconditioned_closure_curvature} on the declared box.

\section{Task-Path Allowance Construction}
\label{app:task_path_bounds}
For a motion factor from Appendix~\ref{app:path_majorants}, the matrix $\boldsymbol X_i=\boldsymbol\xi_i^\wedge$ is its constant generator, $\boldsymbol U_{i,0}$ is its value at the declared center, and $b_{q,j(i)}$ is the corresponding physical-coordinate error allowance from~\eqref{eq:module_allowance_assembly}. Commutation of the center and incremental exponents gives the entrywise factor allowance
\begin{equation}
 \boldsymbol B_{U_i}=|\boldsymbol U_{i,0}|
 \left(\exp(|\boldsymbol X_i|b_{q,j(i)})-\boldsymbol I_4\right).
 \label{eq:task_factor_allowance}
\end{equation}
Expansion of the matrix exponential shows that $|\boldsymbol U_i-\boldsymbol U_{i,0}|\leq\boldsymbol B_{U_i}$. Fixed nominal factors have zero allowance. Repeated application of~\eqref{eq:componentwise_product_allowance} gives the end-point transform allowance $\boldsymbol B_T$ and the corresponding prefix allowances.

For a nonnegative vector $\boldsymbol b\in\R^3$, the matrix $\boldsymbol C(\boldsymbol b)=|[\boldsymbol b]_\times|$ encloses the corresponding skew-matrix perturbation. A prefix $\boldsymbol P=[\rot_P\ \boldsymbol r_P;\boldsymbol0^{\mathsf T}\ 1]$ has rotation and translation allowances $\boldsymbol B_{\mathcal R_P}$ and $\boldsymbol b_{r_P}$ obtained from its transform allowance. Its angular-first adjoint has blocks $[\rot_P,0;[\boldsymbol r_P]_\times\rot_P,\rot_P]$~\cite{lynch2017modern}. The associated entrywise allowance therefore has diagonal blocks $\boldsymbol B_{\mathcal R_P}$, a zero upper-right block, and lower-left block
\[
 |[\boldsymbol r_{P,0}]_\times|\boldsymbol B_{\mathcal R_P}
 +\boldsymbol C(\boldsymbol b_{r_P})(|\rot_{P,0}|+\boldsymbol B_{\mathcal R_P}).
\]
Here $\rot_{P,0}$ and $\boldsymbol r_{P,0}$ are the prefix center values. Applying this construction to $\boldsymbol A_i$ gives the adjoint allowance $\boldsymbol B_{\Ad_{A_i}}$. In physical coordinates, the path-column allowance is $\boldsymbol B_{\eta,j}=\sum_{i:j(i)=j}\boldsymbol B_{\Ad_{A_i}}|\boldsymbol\xi_i|$. Collecting its angular and linear blocks over all $j$ gives the matrices $\boldsymbol B_\Omega$ and $\boldsymbol B_V$. Together with $\boldsymbol b_r=\boldsymbol B_T(1{:}3,4)$, these bound the end-point velocity map by
\begin{equation}
 \boldsymbol B_F=\begin{bmatrix}
 \boldsymbol B_V+|[\boldsymbol r_0]_\times|\boldsymbol B_\Omega+
 \boldsymbol C(\boldsymbol b_r)(|\boldsymbol\Omega_0|+\boldsymbol B_\Omega)\\
 \boldsymbol B_\Omega
 \end{bmatrix}.
 \label{eq:world_task_path_allowance}
\end{equation}
Absent task-path coordinates remain zero columns. The end-point transform bound also supplies the fixed-active static position and geodesic rotation allowances in Corollary~\ref{cor:static_pose}.

\section*{Acknowledgements}
We acknowledge the financial support of the Finnish Ministry of Education and Culture through the Intelligent Work Machines Doctoral Education Pilot Program (IWM VN/3137/2024-OKM-4).

\bibliographystyle{elsarticle-num}
\bibliography{ref}
\end{document}